\documentclass[journal]{IEEEtran}
\usepackage{booktabs}
\usepackage{graphicx}
\usepackage{enumitem}
\usepackage{amssymb} 
\usepackage{titlesec}
\usepackage{authblk}
\usepackage{CJKutf8}
\usepackage{wrapfig}
\usepackage{lipsum}
\usepackage[numbers,sort&compress]{natbib}
\usepackage[table]{xcolor}
\usepackage{color}
\usepackage{multirow}
\usepackage{subcaption}
\ifCLASSINFOpdf
\else
\fi
\usepackage{amsmath} 
\usepackage{threeparttable}
\usepackage{algorithm}
\usepackage{algorithmic}

\begin{document}
%
\title{{Multi-objective Genetic Programming with Multi-view Multi-level Feature for Enhanced Protein Secondary Structure Prediction}}
%
%


\author{%
    Yining Qian, Lijie Su, Meiling Xu, \textit{Member, IEEE} and Xianpeng Wang, \textit{Senior  Member, IEEE}%
    \thanks{This work has been submitted to the IEEE for possible publication. Copyright may be transferred without notice, after which this version may no longer be accessible.}%
    \thanks{This work was supported by the Major Program of the National Natural Science Foundation of China (72192830, 72192831), and the National Natural Science Foundation of China (62176049), and the 111 Project (B16009). (\textit{Corresponding author: Lijie Su. Email: sulijie@ise.neu.edu.cn})}%
    \thanks{Yining Qian is with the National Frontiers Science Center for Industrial Intelligence and Systems Optimization, Shenyang 110819, China, and also with the Key Laboratory of Data Analytics and Optimization for Smart Industry (Northeastern University), Ministry of Education, Shenyang 110819, China.}%
    \thanks{Lijie Su is with the Liaoning Engineering Laboratory of Data Analytics and Optimization for Smart Industry, Shenyang 110819, China.}%
    \thanks{Meiling Xu is with the Liaoning Key Laboratory of Manufacturing System and Logistics Optimization, Shenyang 110819, China.}%
    \thanks{Xianpeng Wang is with the Liaoning Engineering Laboratory of Data Analytics and Optimization for Smart Industry, Shenyang, 110819, China.}%
}

\begin{CJK*}{UTF8}{gbsn}
\maketitle

\begin{abstract}
\textcolor{black}{Predicting protein secondary structure is essential for understanding protein function and advancing drug discovery. However, the intricate sequence-structure relationship poses significant challenges for accurate modeling. To address these, we propose MOGP-MMF, a multi-objective genetic programming framework that reformulates PSSP as an automated optimization task focused on feature selection and fusion. Specifically, MOGP-MMF introduces a multi-view multi-level representation strategy that integrates evolutionary, semantic, and newly introduced structural views to capture the comprehensive protein folding logic. Leveraging an enriched operator set, the framework evolves both linear and nonlinear fusion functions, effectively capturing high-order feature interactions while reducing fusion complexity. To resolve the accuracy-complexity trade-off, an improved multi-objective GP algorithm is developed, incorporating a knowledge transfer mechanism that utilizes prior evolutionary experience to guide the population toward global optima. Extensive experiments across seven benchmark datasets demonstrate that MOGP-MMF surpasses state-of-the-art methods, particularly in Q8 accuracy and structural integrity. Furthermore, MOGP-MMF generates a diverse set of non-dominated solutions, offering flexible model selection schemes for various practical application scenarios. The source code is available on GitHub: https://github.com/qian-ann/MOGP-MMF/tree/main.}
\end{abstract}


\IEEEpeerreviewmaketitle

\section{Introduction}

\begin{figure}[t]
 \centering
  \includegraphics[width=6cm,height=3cm]{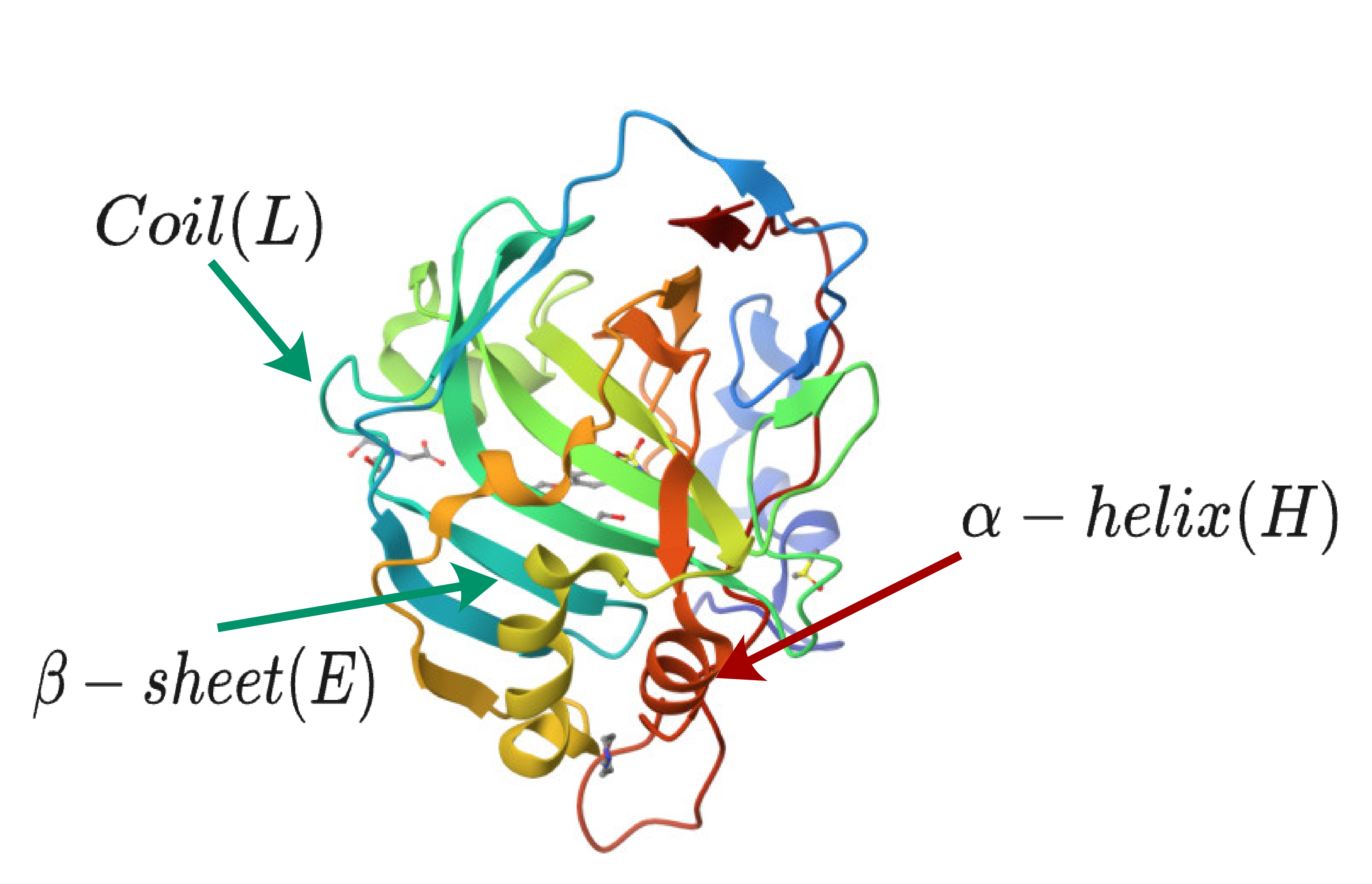}
  \caption{\footnotesize Crystal structure of chimeric carbonic anhydrase VI bound to ethoxzolamide (PDB ID: 6QL2), illustrating the protein secondary structure prediction.}
    \label{PSSP}
\end{figure}

Protein Secondary Structure Prediction (PSSP) classifies each amino acid residue in a protein sequence into specific structural categories, such as $\alpha$-helix, $\beta$-sheet, or coil, based on its local conformation \cite{properties}, as illustrated in Fig. \ref{PSSP}. Accurate PSSP not only imposes critical constraints for three-dimensional structure modeling but also plays a crucial role in understanding protein function and advancing drug discovery \cite{CACE}. Consequently, PSSP remains a core research topic in computational biology \cite{qian}.

Despite extensive research, the complex mapping between protein structure and sequence presents significant challenges in building high-accuracy prediction models \cite{2022deep}. Existing approaches primarily include physical mechanism-based and data-driven models. Physical mechanism-based methods, such as energy minimization models, provide theoretical interpretability but are computationally expensive and inefficient for large-scale data. In contrast, data-driven methods, particularly deep learning techniques, have gained popularity due to their powerful high-dimensional feature extraction capabilities \cite{fold}. For instance, studies such as \textcolor{black}{\cite{NetSurfP-3.0, ShuffleNet_SS, AttSec, DeepPredict}} employed single-view or single-feature extraction techniques, while others, \textcolor{black}{including \cite{Mufold-ss, 2C-BLSTM, DeepACLSTM, BRNN, SADGRU-SS, SERT, TruMPET, MFTrans, PSSP-MFFNet}}, integrated multiple feature processors to enhance performance. Although these methods have shown promising results, they often rely heavily on expert knowledge, requiring extensive fine-tuning and fixed architectures. This limits their adaptability to complex and diverse data, hindering global optimization. Single-feature extraction approaches often fail to capture intricate relationships between protein sequences and structures, resulting in information loss and decreased accuracy. On the other hand, multi-feature learning approaches may lead to inefficiencies, feature redundancy, and the omission of critical features. These challenges are not exclusive to PSSP but are prevalent across various protein-related tasks.

Genetic Programming (GP), an extension of evolutionary algorithms (EAs), is known for its flexible model representation and global search capabilities \cite{SMC1}. GP constructs mathematical expressions to describe relationships between input and output variables, achieving feature learning through adaptive feature selection and integration. Its ability to evolve solutions without predefined structures makes it highly adaptable to diverse problems \cite{2024genetic}. GP has achieved notable success in various fields, such as text classification \cite{text}, path planning \cite{path}, flow-shop scheduling \cite{flow}, and image classification \cite{bi2020,bi2022}. For example, in image classification tasks \cite{EDF,bi2020}, GP employs fixed program structures combined with simple functions for feature extraction and classification. In industrial quality prediction \cite{industry1}, mechanism-based knowledge and data analysis methods (e.g., correlation analysis, feature importance ranking) guide feature construction, with GP applied for feature selection and integration, yielding outstanding predictive performance. However, unlike standardized image or industrial data, protein data is highly complex, making function-based feature extraction challenging. Furthermore, although deep learning effectively extracts complex features, its direct integration into previous GP methods can introduce significant computational overhead \cite{MO4}.

This study proposes a multi-objective GP-based approach, termed MOGP-MMF, to address the challenges of PSSP. The model employs a multi-view multi-level feature extraction strategy to achieve high accuracy and generalization performance through automated feature fusion while mitigating information loss and redundancy. Additionally, MOGP-MMF provides diverse solutions options for various application scenarios. The main contributions of this work are summarized as follows:

\textcolor{black}{(\romannumeral1) This study redefines PSSP as a joint optimization problem of autonomous feature selection and fusion. A novel GP-based approach is introduced, leveraging flexible representation and global search to support adaptive model evolution. This eliminates the dependency on expert-designed models and offers a novel "automated modeling" solution for capturing complex protein sequence-structure relationship.}

\textcolor{black}{(\romannumeral2) A multi-view multi-level feature (MMF) strategy is proposed to alleviate the information bottleneck. MMF integrates evolutionary (HMM/PSSM), semantic (ProtTrans-T5), and structural (SaProt) views to ensure a robust capture of protein folding patterns. Furthermore, an enriched operator set (EOS) is developed to evolve both linear and nonlinear fusion functions. This approach not only captures high-order feature interactions but also reduces structural complexity, resolving the limitations of rigid feature concatenation.}

\textcolor{black}{(\romannumeral3) To balance predictive accuracy and model complexity, we develop an improved multi-objective GP (MOGP) framework featuring a knowledge transfer (KT) mechanism. By leveraging high-performing individuals from single-objective evolution as prior knowledge, this approach effectively avoids local optima while maintaining model simplicity. Consequently, it generates a diverse set of non-dominated solutions, enabling flexible model selection for diverse practical applications.}

\textcolor{black}{(\romannumeral4) Extensive experiments across seven benchmark datasets and six evaluation metrics validate the effectiveness and robustness of MOGP-MMF. The evolved analytical solutions offer transparent insights into optimal feature selection and fusion logic, while high-dimensional visualization of the learned representations provides a profound understanding of the model's superior discriminative power.}

\section{RELATED WORKS}

\subsection{Existing prediction methods for protein secondary structure}

\textcolor{black}{PSSP maps an amino acid sequence $P = \{A_i\}_{i=1}^n$ to a structural sequence $S = f(P) = \{S_i\}_{i=1}^n$. Current research focuses on eight-state models (distinguishing $\alpha, 3_{10}, \pi$ helices, $\beta$-sheets, $\beta$-bridges, turns, loops, and coils), which offer fine-grained structural details but impose significant feature extraction challenges. Given that experimental methods like X-ray crystallography are costly and time-consuming, the field has largely shifted toward data-driven computational approaches \cite{EDF}. Early computational approaches evolved from SVM-based methods \cite{MSBTCN} to deep learning architectures capable of capturing complex dependencies. Hybrid models employing CNNs and RNNs, such as 2C-BLSTM \cite{2C-BLSTM}, DeepACLSTM \cite{DeepACLSTM}, and MUFOLD-SS \cite{Mufold-ss}, became prevalent for modeling local and long-range interactions, while ShuffleNet-SS \cite{ShuffleNet_SS} explored lightweight convolutional efficiency. The subsequent integration of attention mechanisms and Pre-trained Language Models (PLMs) (e.g., NetSurfP-3.0 \cite{NetSurfP-3.0}, AttSec \cite{AttSec}, SADGRU-SS \cite{SADGRU-SS}) further enhanced global semantic representation. Most recently, research has shifted towards multi-view fusion and MSA-free strategies. To maximize information extraction, models like MFTrans \cite{MFTrans} and PSSP-MFFNet \cite{PSSP-MFFNet} utilize Transformer-based architectures to synthesize high-dimensional embeddings, while SERT-StructNet \cite{SERT} refines multi-factor selection via dilated convolutions. Concurrently, to address alignment bottlenecks, MSA-free approaches such as DeepPredict \cite{DeepPredict} and TruMPET \cite{TruMPET} leverage ESM-2 or ESMFold2 embeddings to enable rapid, high-accuracy predictions.}

\textcolor{black}{Despite these advancements, most existing models rely on rigid architectures that lack adaptive feature fusion, leading to suboptimal information integration and feature redundancy. Furthermore, they typically prioritize a single accuracy metric while neglecting model complexity, which may hinder flexible deployment across diverse resource-constrained environments. These limitations necessitate an automated, multi-objective framework, motivating our genetic programming approach.}

\subsection{GP-based methods for PSSP}

The application of GP in PSSP remains relatively unexplored. For example, \cite{2015conference} applied GP to model amino acid sequences using basic clustering techniques for feature learning, offering a simple and cost-effective solution. Similarly, \cite{2004} utilized GP to refine prediction probabilities generated by the PSIPRED neural network by performing local or global averaging operations within fixed windows, leading to improved accuracy. Despite these efforts, existing GP-based methods primarily represent protein sequences from a single perspective and rely on traditional machine learning techniques. They often employ basic linear operators to transform feature matrices and integrate information, which may fail to capture the intricate relationships between protein sequences and structures, ultimately limiting predictive performance. 

In contrast, this study formulates PSSP as an optimization problem focused on feature selection and fusion. A multi-view multi-level feature representation strategy is proposed to address the limitations of current protein feature representation. Additionally, an enriched operator set that combines linear and nonlinear operators is designed to improve the flexibility and effectiveness of feature fusion. To further enhance model applicability, an improved multi-objective strategy is introduced, balancing performance and complexity based on practical application requirements. These enhancements aim to make GP-based methods more robust and adaptable to diverse PSSP tasks.

\section{METHODOLOGY}
\begin{figure*}[htbp]
 \centering
\includegraphics[width=14cm,height=13cm]{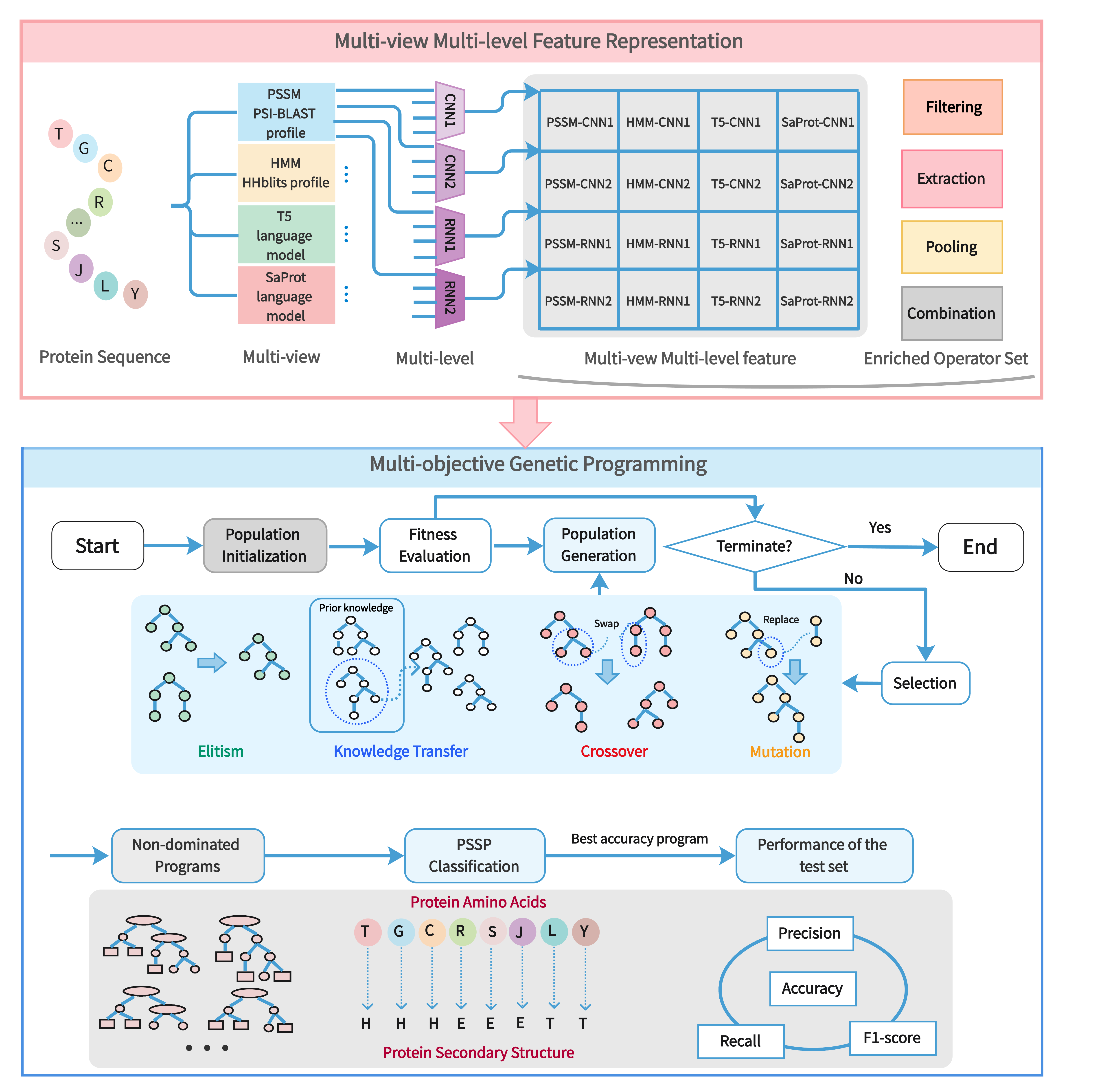}
\caption{\small The overall workflow of MOGP-MMF. It involves extracting multi-view features (PSSM, HMM, T5, SaProt), performing evolutionary feature fusion via an enriched operator set and genetic strategies, and finally selecting the optimal program for performance evaluation.}
 \label{f002}
\end{figure*}


\subsection{Framework of MOGP-MMF}

The overall framework of MOGP-MMF, illustrated in Fig.~\ref{f002}, integrates three core components: the multi-view multi-level feature representation (MMF), the enriched operator set (EOS), and the improved multi-objective genetic programming (MOGP) engine. Their specific roles are detailed below.

The MMF module extracts features from raw protein sequences to address the limitations of existing approaches restricted to narrow feature scopes. It integrates diverse perspectives, specifically evolutionary, semantic, and structure-aware views, while employing both local and global extractors. This results in a robust multi-level representation that effectively captures the intricate mapping between protein sequences and their structures.

The EOS component introduces a set of both linear and nonlinear operators to address the limitations of solely using linear functions. It enhances feature fusion flexibility by incorporating operations such as filtering, feature extraction, pooling, and concatenation. These operations enable the capture of intricate feature interactions, improving representation quality.

The MOGP component builds the protein secondary structure prediction model through a series of evolutionary processes, including population initialization, fitness evaluation, evolutionary operations, and non-dominated solution generation. Fitness evaluation considers both prediction accuracy and model complexity, with adjustable cost weights for views and feature usage to enhance applicability. Evolutionary operations—such as elitism, transfer knowledge, crossover, and mutation—optimize the population and explore diverse feature fusion strategies. The resulting non-dominated solution set provides adaptable models for various applications. To comprehensively evaluate the model’s performance, metrics such as precision, recall, and F1-score are utilized to ensure robustness and reliability.


\subsection{MMF Method}

The MMF method comprises two main stages: multi-view information representation and multi-level feature extraction.

\begin{figure}[htbp]
 \centering
  \includegraphics[width=9cm,height=12cm]{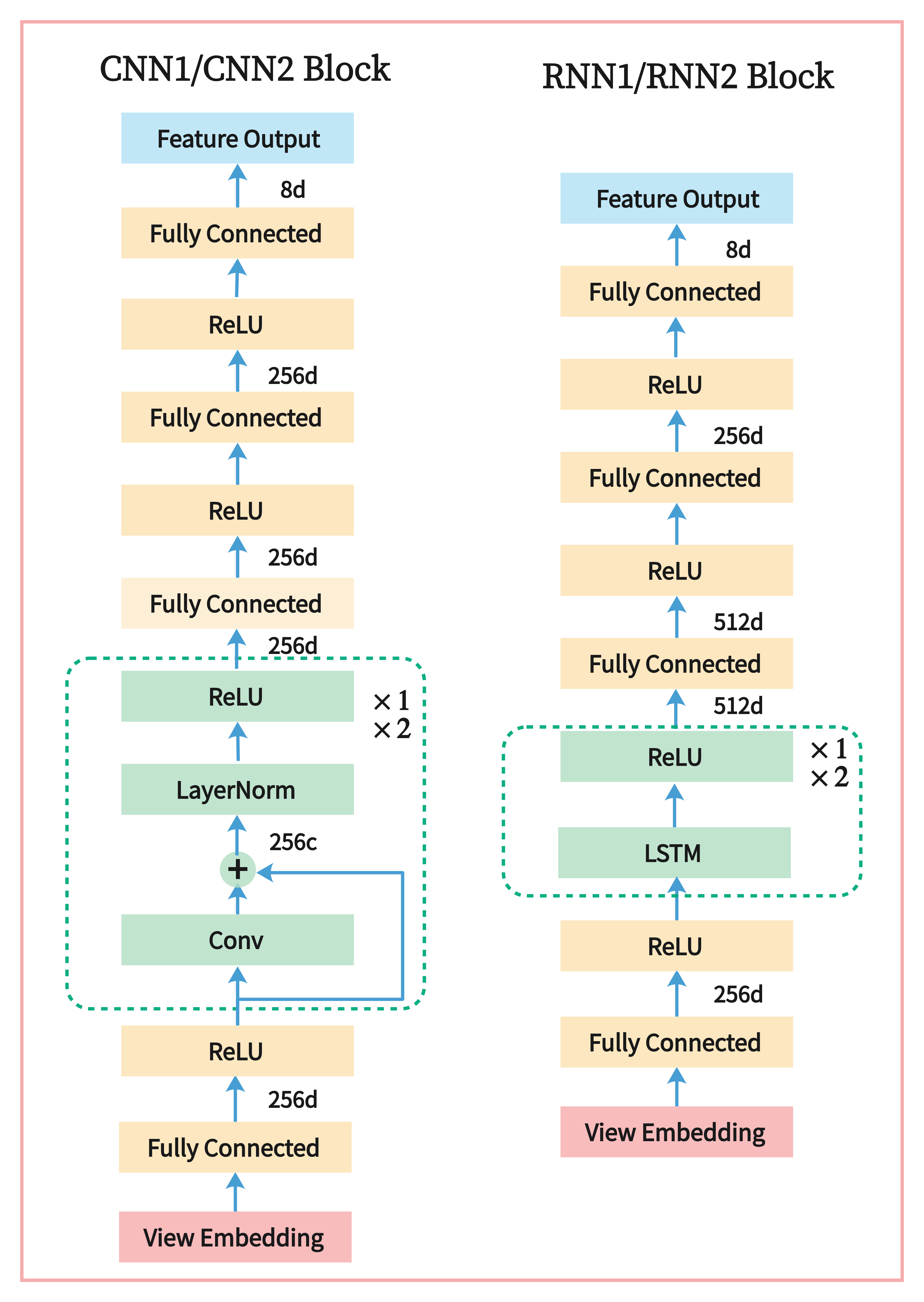}
  \caption{\small Architecture of multi-level feature extractors.}
    \label{feature}
\end{figure}

\subsubsection{Multi-view protein representation}

\textcolor{black}{Multi-view representation aims to capture diverse and complementary biological insights by integrating information from multiple dimensions. To construct a holistic and robust protein representation, this study selects four distinct feature sets categorized into three complementary views: the \textbf{Evolutionary View} (PSSM \cite{BRNN} and HMM \cite{HHblits}), the \textbf{Semantic View} (ProtTrans-T5) \cite{T5}, and the \textbf{Structure-aware View} (SaProt) \cite{SaProt}.}

\textbf{Evolutionary View (PSSM \& HMM)}:
This view integrates Position-Specific Scoring Matrices (PSSM) \cite{BRNN} and Hidden Markov Model (HMM) profiles \cite{HHblits} to capture evolutionary conservation and mutation patterns across homologous sequences. PSSM analyzes amino acid frequencies to highlight functionally stable regions, generating an $L\times 21$ matrix. Complementarily, HMM profiles model transition probabilities among hidden states to reveal statistical alignment relationships, providing an $L\times 20$ matrix. Together, they form a robust baseline of evolutionary information.

\textbf{Semantic View (ProtTrans-T5)}:
ProtTrans-T5 \cite{T5} serves as the semantic backbone. As a transformer-based protein language model, it treats sequences as biological sentences, applying NLP techniques to extract high-level contextual and semantic dependencies often missed by statistical models. It generates a high-dimensional feature embedding of size $L\times 1024$.

\textcolor{black}{\textbf{Structure-aware View (SaProt)}:  
SaProt \cite{SaProt} serves as the structure-aware view. Unlike standard sequence models, SaProt is pre-trained on structure-sequence pairs using the Foldseek structure alphabet. This unique mechanism allows it to embed implicit 3D structural constraints and folding patterns directly into the feature space. It produces a compact embedding of size $L\times 480$, effectively capturing essential structural information to complement the semantic features of ProtTrans-T5.}

\subsubsection{Multi-level feature extraction}

Capturing multi-level features is crucial for PSSP, as secondary structures are influenced by both local and long-range interactions among residues. To achieve this, the study employs four distinct extractors (CNN1, CNN2, RNN1, RNN2), as illustrated in Figure \ref{feature} and Algorithm \ref{algorithm1}.

All feature extractors follow a unified process that includes input preparation, feature extraction, dimensionality reduction, and output generation. Each view from the training dataset $D$ and validation dataset $\hat{D}$ is independently processed using CNN1, CNN2, RNN1, and RNN2. The input sequence is concatenated with a 22-dimensional one-hot encoding vector (Step 1) and passed through a fully connected layer (Step 2) to generate a 256-dimensional vector. This vector is subsequently fed into the CNN or RNN layers (Step 3) for feature extraction, followed by three fully connected layers (Step 4) for further processing. The extracted features undergo dimensionality reduction, and the final 8-dimensional output from Step 5 is concatenated with the previous 256-dimensional vector, forming a 264-dimensional final representation (Step 6). 

Ultimately, 16 feature representations are generated. All features are normalized using $L2$ normalization to ensure consistency across different feature types and enable unbiased comparisons. This study focuses on widely used views and feature extractors while maintaining flexibility to integrate additional sources of information.


\begin{algorithm}
\caption{Flowchart of Multi-level Feature Extraction}
\label{algorithm1}
\textbf{Input:} Training dataset with multi-views \\
\hspace*{3em} $D =\{(V_{HMM},Y), (V_{PSSM},Y),(V_{T5},Y),(V_{Sa},Y)\}$;\\
\hspace*{3em} Validation dataset with multi-views \\
\hspace*{3em} $\hat{D}= \{(\hat{V}_{HMM},\hat{Y}), (\hat{V}_{PSSM},\hat{Y}),(\hat{V}_{T5},\hat{Y}),(\hat{V}_{Sa},\hat{Y})\}$.\\
\textbf{Output:} Multi-feature set $F = \{F_i\}_1^{16}$.\\
\begin{algorithmic}[1]
\STATE Concatenate the view information with a 22-dimensional one-hot encoding to obtain the view embedding;
\STATE Input view embedding to a fully connected layer with ReLU, output a 256-dimensional vector;
\STATE Input the obtained vector to:\\
 \quad - Option 1: One CNN layer (CNN1);\\
 \quad - Option 2: Two CNN layers (CNN2);\\
 \quad - Option 3: One BiLSTM layer (RNN1);\\
 \quad - Option 4: Two BiLSTM layers (RNN2);\\
\STATE Input the obtained vector to:\\
\quad - Case 1 (CNN1 or CNN2): Three fully connected layers with ReLU activation, resulting in outputs of 256, 256, and 8 dimensions, respectively;\\
\quad - Case 2 (RNN1 or RNN2): Three fully connected layers with ReLU activation, resulting in outputs of 512, 256, and 8 dimensions, respectively.
\STATE Apply Softmax function, output predictions;
\STATE \textbf{return} $F_i$ obtained by concatenating the output vectors of the last (8-dimensional) and the second-to-last (256-dimensional) fully connected layer, yielding a 264-dimensional vector.
\end{algorithmic}
\end{algorithm}

\subsection{EOS Method}

To capture complex interactions between features beyond traditional linear limits, we propose an Enriched Operator Set (EOS) comprising both linear and nonlinear functions tailored for feature fusion (see Table \ref{functions}). The EOS consists of four functional categories:\begin{itemize}\item \textbf{Filtering:} Includes linear weighted operations ($W_{Add}$, $W_{Sub}$) for feature alignment, and nonlinear transformations ($Mul$, $GRT$, $Sqrt$, $Log$, $Exp$, $ReLU$) to enhance feature expressiveness and diversity.\item \textbf{Feature Extraction:} $LoGF$ (Laplacian of Gaussian) highlights structural details via smoothing and edge detection, while $FFT$ extracts frequency-domain patterns to identify periodic structures.\item \textbf{Pooling:} $MaxP$ reduces dimensionality via sliding windows, preserving critical local features.\item \textbf{Concatenation:} $Root1–3$ enable the flexible merging of 1 to 3 feature branches, supporting diverse tree architectures.\end{itemize}

Additionally, Ephemeral Random Constants are introduced to dynamically optimize operator parameters (Table \ref{constant}). These constants, such as the smoothing factor $\sigma$ in $LoGF$, the mode choice in $FFT$, and weights $n$ in linear functions, are initialized randomly and refined through evolutionary mutation, thereby enhancing the model's adaptability.

\begin{table}[ht]
\caption{Functions in EOS}
\centering
\setlength{\tabcolsep}{6pt}
\renewcommand{\arraystretch}{1}
\begin{tabular}{p{1.2cm}p{1.5cm}p{5cm}}
\toprule
\textbf{Function} & \textbf{Type}& \textbf{Description} \\
\midrule
\textit{$W\_Add$} & linear&Add two weighted features\\ 
\textit{$W\_Sub$} & linear&Subtract two weighted features\\
\textit{Mul} & non-linear &Multiply filter \\
\textit{GRT} & non-linear&Select maximum value between two features \\ 
\textit{Sqrt} & non-linear&Sqrt an feature \\
\textit{Log} & non-linear&Logarithm function \\
\textit{Exp} & non-linear&Exponential function  \\
\textit{ReLU} & non-linear&The rectified linear unit \\
\textit{LoGF} &non-linear&Combine Gaussian and Laplacian filtering \\
\textit{FFT} &non-linear& One-dimensional Fourier transform \\
\textit{MaxP} & non-linear&Max pooling  \\ 
\textit{Root 1/2/3} & non-linear&Concatenate vectors to a vector\\
\bottomrule
\end{tabular}
\label{functions}
\end{table}

\begin{table}[ht]
\caption{Ephemeral random constants in EOS}
\centering
\setlength{\tabcolsep}{5pt}
\renewcommand{\arraystretch}{1}
\begin{tabular}{lp{1.5cm}p{5cm}}
\toprule
\textbf{Constant} & \textbf{Type} & \textbf{Description} \\
\midrule
\textit{$n$} & Integer & The parameters for the \textit{$W\_Add$} and \textit{$W\_Sub$} functions. They are randomly generated in the range of [1,10).\\ 
\textit{$\sigma$}& Integer & The convolution kernel size of \textit{$LoGF$} is with values ranging from 1 to 4.\\
\textit{$FFTmode$} &Integer& The transform dimension for \textit{$FFT$} with possible values of 1.\\
\bottomrule
\end{tabular}
\label{constant}
\end{table}

\subsection{MOGP Method}

The MOGP approach optimizes a population of candidate models for PSSP using evolutionary algorithms. Initially, the population $\mathcal{P}_g$ is generated through uniform random sampling, where each program is evaluated using Eq.(\ref{eq1}) and Eq.(\ref{eq2}) (lines 1--3). The optimization process continues until the maximum number of generations, $G_{max}$, is reached (lines 5--11). First, an elite population $I$, consisting of $\rho N$ top-performing individuals, is selected using the elitism operator (line 5). A subset $S$ is then chosen from $\mathcal{P}_g$ via tournament selection based on prediction accuracy (line 6).

Next, offspring individuals are generated from $S$ using subtree crossover with knowledge transfer and subtree mutation operators (line 7). Each program in the offspring set $\mathcal{O}$ is evaluated using Eq.(\ref{eq1}) and Eq.(\ref{eq2}) (line 8). The final offspring $\mathcal{o}_g$ is selected from $\mathcal{Q}$ using fast non-dominated sorting based on constraint dominance (line 9). The next-generation population $\mathcal{P}_{g+1}$ is then formed by combining the offspring set $\mathcal{Q}_g$ with the elite population $I$ (line 10).

This study focuses on tree-based GP, where programs are represented using tree structures. In the proposed MOGP approach, both prediction accuracy and model complexity are considered to balance model performance with structural efficiency. Complexity is constrained by limiting the number of features, views, and nodes in the generated programs. To further improve prediction accuracy, a knowledge transfer technique is incorporated during the offspring generation process. The following subsections discuss the key steps of MOGP: program evaluation and knowledge transfer.

\subsubsection{Program Evaluation}
The fitness of each individual is assessed based on two objectives: predictive accuracy and model complexity, inspired by \cite{industry2, industry3}. Accuracy measures the proportion of correctly predicted residues, while complexity minimizes redundancy by reducing unnecessary features. The non-dominated sorting method is employed to evaluate solution quality.

\begin{itemize}
\item {\it \textbf{Accuracy}}: For a given individual $p_k$, predictions $\hat{Y}$ are generated using the {\it Decoding} component. Since PSSP is a classification task, accuracy is calculated as follows:
\begin{equation}\label{eq1}
\begin{aligned}
Maximize: \mathrm{Q}_a = \frac{\sum_{i=1}^{m} N_{i}}{N_\textsl{res}} \times 100\%
\end{aligned}
\end{equation}
where \textsl{m} represents the number of structural states, $N_\textsl{res}$ is the total number of residues, and $N_{i}$ denotes the correctly predicted residues for state ${i}$. Accuracy is determined by comparing the predicted values $\hat{Y}$ with the true labels $Y$.

\item {\it \textbf{Complexity}}: The second objective aims to minimize feature redundancy by reducing unnecessary elements in the fusion tree. For an individual $p_k = [f_k, o_k, w_k]$, feature complexity is defined as:
\begin{equation}\label{eq2}
Minimize: \mathrm{Q}_c = \text{F}(f_k)
+10\text{V}(f_k)+0.1\text{L}(f_k)
\end{equation}
where $\text{F}(f_k)$ represents the number of features, $\text{V}(f_k)$ denotes the number of views, and $\text{L}(f_k)$ represents the number of nodes in the tree. Given that the high cost of obtaining protein view information, reducing the number of views is prioritized in complexity minimization. Additionally, limiting the number of nodes helps control the overall model size.
\end{itemize}

Accuracy is the primary objective in Algorithm \ref{alg2}, guiding the selection process (line 6), while complexity serves to control model efficiency.

\subsubsection{Knowledge Transfer}
In evolutionary algorithms, individuals tend to converge to similar structures over successive generations, which may result in premature convergence to local optima \cite{SMC2}. Additionally, in multi-objective optimization, balancing multiple objectives often results in suboptimal performance for individual objectives. To address these challenges, prior knowledge is incorporated into the evolutionary process to maintain population diversity and enhance optimization performance.

Prior knowledge consists of individuals generated by a single-objective GP (SOGP) algorithm. The SOGP algorithm, which excludes the complexity constraint and knowledge transfer mechanism, is executed five times, with ten individuals selected from each final population, resulting in a total of 50 individuals. As outlined in lines 5–9 of Algorithm \ref{alg3}, these prior individuals ($\mathcal{P}_p$) are integrated into the reproduction process. The control parameter $\mu$ is used to adaptively adjust the source of mating programs. When the random number generated in $[0, 1]$ is less than $\mu$, a mating program is selected from the selection population $S$; otherwise, a priori program will be randomly selected as a mating program from $\mathcal{P}_p$. 

To prevent over-reliance on prior knowledge, which could lead to local optima, the parameter $\mu$ is dynamically updated during the evolutionary process using the following equation (line 12 in Algorithm \ref{alg2}):
\begin{equation}\label{mueq}
\begin{aligned}
\mu=\frac{1}{2}+\frac{N_{P}}{N_{S}+N_{P}}
\end{aligned}
\end{equation}
where $N_{S}$ and $N_{P}$ represent the selection counts from $S$ and $\mathcal{P}_p$ (Algorithm \ref{alg3}-lines 6 and 8) in the previous generation, respectively. Initially, $N_{S}=N_{P}=0$. This equation ensures that $\mu\ge\frac{1}{2}$, meaning the contribution of prior knowledge to the evolution process is limited and gradually adjusted based on prior selection frequencies.

\begin{algorithm}
\caption{Framework of MOGP}
\label{alg2}
\renewcommand{\arraystretch}{0.8}
\small
\textbf{Input:} Population size $N$; maximal generation $G_{max}$;\\
\hspace*{3em} enriched operator set $O$; MMF set $F$; ;\\
\hspace*{3em} priori knowledge set $\mathcal{P}_p$; labels $Y$.\\
\textbf{Output:} Nondominated program set $\mathcal{P}^{*}$. \\

\begin{algorithmic}[1]
    \STATE $\mathcal{P}_0$ $\leftarrow$ Initialize the population using the ramped half-and-half method according to the new program structure, the new function and terminal sets;
    \STATE $g \leftarrow 0$ Initialize generation counter
    \STATE $\text{GET\_FITNESS}(\mathcal{P}_0,F,Y)$ $\leftarrow$ Evaluate the individuals in $\mathcal{P}_0$ using Eq.\ref{eq1} and Eq.\ref{eq2};
\WHILE{$g < G_{\text{max}}$}
    \STATE $I$ $\leftarrow$ The best $\rho N$ individuals of $\mathcal{P}_g$ using elitism operator;
    \STATE $S$ $\leftarrow$ $N$ Individuals selected from $\mathcal{P}_g$ using tournament selection according to the first fitness (prediction accuracy);
    \STATE $\mathcal{O}$ $\leftarrow$ Generate the offspring population using reproduction with knowledge transfer strategy shown in Algorithm \ref{alg3};
    \STATE $\text{GET\_FITNESS}(\mathcal{P}_g,F,Y)$ $\leftarrow$ Evaluate the fitness of the individuals in $\mathcal{O}$；
    \STATE $\mathcal{O}_{g}$ $\leftarrow$ select $(1-\rho)N$ individuals from $\mathcal{O}$ based on non-dominated sorting (like NSGA-II);
    \STATE $\mathcal{P}_{g+1}$ $\leftarrow$ $\mathcal{O}_{g}\cup I$;
    \STATE $g \leftarrow g+1$;
    \STATE $\mu$ $\leftarrow$ Update $\mu$ according to equation \ref{mueq};
\ENDWHILE
    \STATE ${\mathcal{P}^{*}} \leftarrow$ Select 
    non-dominated programs in $\mathcal{P}_{G_{max}}$ as the candidate programs. 
\end{algorithmic}
\end{algorithm}

\begin{algorithm}
\caption{Offspring generation strategy}
\renewcommand{\arraystretch}{0.8}
\label{alg3}
\small
\textbf{Input:} Priori knowledge set $\mathcal{P}_p$;
selection population $S$;\\ 
\hspace*{3em} population size $N$; crossover probability $P_{cx}$; \\ 
\hspace*{3em} priori knowledge selection probability $\mu$.\\
\textbf{Output:} Offspring population $O$.\\
\begin{algorithmic}[1]
    \STATE $\mathcal{O}$ $\leftarrow \emptyset$;
\FOR{$k=1,~k < N,~k++$} 
    \STATE $p_1\leftarrow$ Select the $k$th individual in $\mathcal{S}$ as a mating program;
    \IF{rand()$<P_{cx}$}
    \IF{rand()$<\mu$}
        \STATE $p_2\leftarrow$ Select the $(k+1)$th individual in $\mathcal{S}$ as a mating program;
    \ELSE
        \STATE $p_2\leftarrow$ Randomly select a mating program from $\mathcal{P}_p$;
    \ENDIF
        \STATE $p\leftarrow$ SubtreeCrossover$(p_1,p_2)$;
    \ELSE
        \STATE $p\leftarrow$ SubtreeMutation$(p_1)$;
    \ENDIF
    
    \STATE $\mathcal{O}\leftarrow\mathcal{O}\cup\{p\}$.
\ENDFOR
\end{algorithmic}
\end{algorithm}

\subsection{Test Process} During the evolutionary process of MOGP-MMF, a least squares classifier replaces computationally expensive gradient-based neural networks. This substitution significantly accelerates the evaluation process while maintaining performance, given that deep learning is already employed during feature extraction. The best-performing individual is applied to various test sets to comprehensively assess its accuracy and generalization ability.

\section{Experiment Design}

\subsection{Datasets}

We employed a comprehensive multi-tier dataset protocol consisting of one large-scale training set and six independent benchmark test sets.

\textbf{Training and Validation:} The model is trained on the CB6133 dataset \cite{CB513}, a large-scale benchmark derived from the Protein Data Bank (PDB) and curated via SCOP. It contains 6133 non-redundant sequences, strictly partitioned into training (5600 sequences), testing (277 sequences), and validation (256 sequences) subsets to prevent information leakage.

\textcolor{black}{\textbf{Independent Testing:} For objective performance assessment, we utilized the CB513 dataset \cite{CB513} as the standard reference benchmark. Crucially, to strictly evaluate the model's capability to generalize to novel and increasingly difficult protein folds, we extended the testing suite to include five successive datasets from the Critical Assessment of Structure Prediction (CASP) experiments: CASP10 \cite{CASP10}, CASP11 \cite{CASP11}, CASP12 \cite{CASP12}, CASP13 \cite{CASP13}, and CASP14 \cite{CASP14}. This specifically includes the most recent challenges (CASP12--14), which represent harder targets with lower sequence homology, providing a "stress test" for the model's adaptability to modern protein structures.}

\textbf{Preprocessing:} All sequences were normalized to a fixed length of 700 residues. Sequences shorter than 700 were zero-padded, while longer ones were processed via overlapping slicing to ensure consistent input dimensions.

\subsection{Parameter Settings}
The model is implemented using a single NVIDIA GeForce RTX 4090 GPU with 24GB of memory, and the PyTorch deep learning framework. The Adam optimizer is employed to facilitate effective convergence. Hyperparameters for both the multi-feature extraction and GP-based deep fusion phases are determined through empirical testing and established best practices, ensuring optimal performance and model stability. Table \ref{hyperparameters} presents the key hyperparameters, along with their abbreviations and specific configurations.

\begin{table}[ht]
\caption{Hyperparameters used in GP-MMF}
\centering
\setlength{\tabcolsep}{10pt}
\renewcommand{\arraystretch}{1}
\begin{tabular}{lp{2.8cm}}
\toprule
\textbf{Hyperparameter} & \textbf{Setting} \\
\toprule
Number of epochs & 100 \\
Dropout rate  & 0.2 \\
Batch size  & 64 \\
Maximum sequence length & 700 \\
Number of labels & 8 \\
Kernel size (CNNs)& 9 \\
Padding length (CNNs)& 4 \\
Number of channels (CNNs)& 256 \\
Hidden layer size (RNNs) & 256 \\
Fully connected layers (CNNs) & 256, 256, 8 \\
Fully connected layers (RNNs)& 512, 256, 8 \\
Maximum number of generations (GP) & 50 \\
Population size (GP) & 200 \\
Elitism rate (GP) & 0.05 \\
Mutation rate (GP) & 0.2 \\
Crossover rate (GP) & 0.8\\
Tree depth at initialization step (GP) & [2, 6]\\
Maximum tree depth (GP) & 8\\
Random seeds (GP) & 100 \\
\bottomrule
\end{tabular}
\label{hyperparameters}
\end{table}





\subsection{Evaluation}
\textcolor{black}{Performance is evaluated using six metrics: Accuracy (Q8), Precision, Recall, F1-score, Matthews Correlation Coefficient (MCC), and Segment Overlap (Sov).}

Accuracy, defined in Eq.(\ref{eq1}), serves as the primary objective. To address class imbalance and assess robustness, we utilize Precision, Recall, and F1-score, defined as:

\begin{equation}\label{eq3}
\small
\text{Precision} = \frac{TP}{TP + FP}
\end{equation}

\begin{equation}\label{eq4}
\small
\text{Recall} = \frac{TP}{TP + FN}
\end{equation}

\begin{equation}\label{eq5}
\small
\text{F1-score} = \frac{2 \cdot \text{Precision} \cdot \text{Recall}}{\text{Precision} + \text{Recall}}
\end{equation}
where TP, FP, FN, and TN represent true positives, false positives, false negatives, and true negatives, respectively.

\textcolor{black}{Furthermore, we introduce MCC (Eq.~\ref{eq6}) to rigorously handle data imbalance using the full confusion matrix, and Sov (Eq.~\ref{eq7}) to evaluate the structural integrity of predicted segments rather than individual residues:}

\textcolor{black}{\begin{equation}\label{eq6}
\small
\text{MCC} = \frac{(TP \cdot TN) - (FP \cdot FN)}{\sqrt{(TP+FP)(TP+FN)(TN+FP)(TN+FN)}}
\end{equation}}

\textcolor{black}{
\begin{equation}\label{eq7}
\small
\text{Sov} = \frac{100}{N} \sum_{s \in S} \left[ \frac{min(s_o, s_p) + \delta(s_o, s_p)}{max(s_o, s_p)} \cdot L(s_o) \right]
\end{equation}}
\textcolor{black}{where $N$ is the total residues; $s_o$/$s_p$ denote observed/predicted segments; $min$ and $max$ denote the segment intersection and union lengths; $L(s_o)$ is the length of $s_o$; and $\delta$ is the tolerance term.}

\section{Results}
This section presents a comprehensive evaluation of the proposed MOGP-MMF model through an in-depth analysis of experimental results.

\begin{table*}[htbp]
\centering
\caption{\textcolor{black}{Performance comparison of MOGP-MMF with state-of-the-art methods on benchmark datasets. The best results are highlighted in bold.}}
\label{tab:performance_comparison}
\small 
\setlength{\aboverulesep}{0pt} 
\renewcommand{\arraystretch}{0.9} 
\begin{tabular*}{\textwidth}{@{\extracolsep{\fill}}llcccccc}
\toprule
Dataset & Method & Q8 & Precision & Recall & F1-score & \textcolor{black}{MCC} & \textcolor{black}{Sov} \\
\midrule
\multirow{10}{*}{CB6133} 
& NetSurfP-3.0 & 0.765 & 0.771 & 0.765 & 0.756 & 0.711 & 76.2 \\
& AttSec & 0.783 & 0.774 & 0.783 & 0.773 & 0.730 & 78.9 \\
& Mufold-SS & 0.711 & 0.696 & 0.711 & 0.685 & 0.627 & 71.7 \\
& ShuffleNet\_SS & 0.735 & 0.723 & 0.735 & 0.725 & 0.667 & 75.1 \\
& DeepACLSTM & 0.733 & 0.720 & 0.733 & 0.716 & 0.662 & 73.8 \\
& 2C-BLSTM & 0.735 & 0.720 & 0.735 & 0.721 & 0.665 & 73.8 \\
& SADGRU-SS & 0.748 & 0.738 & 0.748 & 0.741 & 0.688 & 76.1 \\
& \textcolor{black}{DeepPredict} & 0.798 & 0.759 & 0.798 & 0.773 & 0.735 & 78.9 \\
& \textcolor{black}{TruMPET} & 0.796 & \textbf{0.821} & 0.789 & 0.785 & \textbf{0.759} & 80.5 \\
& \textbf{MOGP-MMF (ours)} & \textbf{0.800} & 0.793 & \textbf{0.800} & \textbf{0.793} & 0.752 & \textbf{80.6} \\
\midrule
\multirow{10}{*}{CB513} 
& NetSurfP-3.0 & 0.702 & 0.706 & 0.702 & 0.688 & 0.631 & 69.2 \\
& AttSec & 0.749 & 0.738 & 0.749 & 0.736 & 0.686 & 75.1 \\
& Mufold-SS & 0.676 & 0.659 & 0.676 & 0.645 & 0.582 & 67.5 \\
& ShuffleNet\_SS & 0.714 & 0.698 & 0.714 & 0.700 & 0.641 & 72.3 \\
& DeepACLSTM & 0.703 & 0.687 & 0.703 & 0.682 & 0.623 & 70.2 \\
& 2C-BLSTM & 0.704 & 0.686 & 0.704 & 0.685 & 0.626 & 70.1 \\
& SADGRU-SS & 0.724 & 0.710 & 0.724 & 0.713 & 0.657 & 73.3 \\
& \textcolor{black}{DeepPredict} & 0.758 & 0.721 & 0.758 & 0.731 & 0.685 & 74.3 \\
& \textcolor{black}{TruMPET} & 0.768 & \textbf{0.763} & \textbf{0.778} & \textbf{0.760} & 0.715 & 76.6 \\
& \textbf{MOGP-MMF (ours)} & \textbf{0.770} & 0.759 & 0.770 & 0.758 & \textbf{0.716} & \textbf{77.1} \\
\midrule
\multirow{10}{*}{CASP10} 
& NetSurfP-3.0 & 0.761 & 0.763 & 0.761 & 0.750 & 0.705 & 75.3 \\
& AttSec & 0.787 & 0.778 & 0.787 & 0.779 & 0.735 & 79.1 \\
& Mufold-SS & 0.707 & 0.687 & 0.707 & 0.681 & 0.621 & 71.6 \\
& ShuffleNet\_SS & 0.734 & 0.732 & 0.734 & 0.729 & 0.673 & 75.1 \\
& DeepACLSTM & 0.734 & 0.721 & 0.734 & 0.719 & 0.664 & 73.8 \\
& 2C-BLSTM & 0.732 & 0.713 & 0.732 & 0.716 & 0.660 & 72.5 \\
& SADGRU-SS & 0.748 & 0.741 & 0.748 & 0.742 & 0.690 & 76.0 \\
& \textcolor{black}{DeepPredict} & 0.793 & 0.756 & 0.793 & 0.768 & 0.728 & 78.3 \\
& \textcolor{black}{TruMPET} & 0.785 & 0.790 & 0.785 & 0.782 & 0.741 & 78.3 \\
& \textbf{MOGP-MMF (ours)} & \textbf{0.805} & \textbf{0.796} & \textbf{0.805} & \textbf{0.798} & \textbf{0.758} & \textbf{80.7} \\
\midrule
\multirow{10}{*}{CASP11} 
& NetSurfP-3.0 & 0.727 & 0.733 & 0.727 & 0.715 & 0.662 & 71.5 \\
& AttSec & 0.754 & 0.741 & 0.754 & 0.742 & 0.691 & 75.7 \\
& Mufold-SS & 0.696 & 0.678 & 0.696 & 0.669 & 0.607 & 69.7 \\
& ShuffleNet\_SS & 0.711 & 0.709 & 0.711 & 0.705 & 0.643 & 72.6 \\
& DeepACLSTM & 0.715 & 0.699 & 0.715 & 0.697 & 0.638 & 71.6 \\
& 2C-BLSTM & 0.711 & 0.691 & 0.711 & 0.693 & 0.632 & 70.2 \\
& SADGRU-SS & 0.726 & 0.715 & 0.726 & 0.718 & 0.662 & 73.5 \\
& \textcolor{black}{DeepPredict} & 0.766 & 0.730 & 0.766 & 0.741 & 0.695 & 75.6 \\
& \textcolor{black}{TruMPET} & 0.761 & \textbf{0.766} & 0.761 & 0.755 & 0.708 & 74.8 \\
& \textbf{MOGP-MMF (ours)} & \textbf{0.776} & 0.763 & \textbf{0.776} & \textbf{0.766} & \textbf{0.720} & \textbf{77.8} \\
\midrule
\multirow{10}{*}{CASP12} 
& NetSurfP-3.0 & 0.733 & 0.720 & 0.733 & 0.716 & 0.652 & 73.4 \\
& AttSec & 0.732 & 0.717 & 0.732 & 0.719 & 0.653 & 75.4 \\
& Mufold-SS & 0.634 & 0.609 & 0.634 & 0.599 & 0.508 & 64.8 \\
& ShuffleNet\_SS & 0.651 & 0.644 & 0.651 & 0.644 & 0.550 & 69.5 \\
& DeepACLSTM & 0.673 & 0.648 & 0.673 & 0.649 & 0.567 & 68.4 \\
& 2C-BLSTM & 0.667 & 0.641 & 0.667 & 0.640 & 0.561 & 66.1 \\
& SADGRU-SS & 0.699 & 0.690 & 0.699 & 0.688 & 0.618 & 73.3 \\
& \textcolor{black}{DeepPredict} & 0.746 & 0.712 & 0.746 & 0.722 & 0.660 & 74.1 \\
& \textcolor{black}{TruMPET} & 0.730 & \textbf{0.742} & 0.730 & 0.721 & 0.665 & 72.7 \\
& \textbf{MOGP-MMF (ours)} & \textbf{0.746} & 0.731 & \textbf{0.746} & \textbf{0.733} & \textbf{0.672} & \textbf{76.3} \\
\midrule
\multirow{10}{*}{CASP13} 
& NetSurfP-3.0 & 0.718 & 0.705 & 0.718 & 0.698 & 0.631 & 70.6 \\
& AttSec & 0.707 & 0.691 & 0.707 & 0.690 & 0.620 & 72.7 \\
& Mufold-SS & 0.630 & 0.608 & 0.630 & 0.589 & 0.504 & 64.6 \\
& ShuffleNet\_SS & 0.655 & 0.637 & 0.655 & 0.642 & 0.553 & 69.0 \\
& DeepACLSTM & 0.655 & 0.631 & 0.655 & 0.627 & 0.545 & 66.4 \\
& 2C-BLSTM & 0.648 & 0.625 & 0.648 & 0.618 & 0.540 & 64.2 \\
& SADGRU-SS & 0.676 & 0.667 & 0.676 & 0.661 & 0.588 & 69.7 \\
& \textcolor{black}{DeepPredict} & 0.720 & 0.701 & 0.726 & 0.689 & 0.648 & 74.2 \\
& \textcolor{black}{TruMPET} & 0.704 & \textbf{0.725} & 0.704 & 0.693 & 0.639 & 71.2 \\
& \textbf{MOGP-MMF (ours)} & \textbf{0.721} & 0.706 & \textbf{0.721} & \textbf{0.704} & \textbf{0.650} & \textbf{74.6} \\
\midrule
\multirow{10}{*}{CASP14} 
& NetSurfP-3.0 & 0.692 & 0.673 & 0.692 & 0.666 & 0.588 & 67.3 \\
& AttSec & 0.674 & 0.648 & 0.674 & 0.653 & 0.568 & 70.4 \\
& Mufold-SS & 0.621 & 0.599 & 0.621 & 0.582 & 0.488 & 61.1 \\
& ShuffleNet\_SS & 0.640 & 0.621 & 0.640 & 0.626 & 0.528 & 67.7 \\
& DeepACLSTM & 0.648 & 0.615 & 0.648 & 0.617 & 0.526 & 63.1 \\
& 2C-BLSTM & 0.629 & 0.600 & 0.629 & 0.595 & 0.506 & 60.3 \\
& SADGRU-SS & 0.665 & 0.647 & 0.665 & 0.649 & 0.565 & 68.2 \\
& \textcolor{black}{DeepPredict} & 0.680 & 0.667 & 0.689 & 0.661 & 0.586 & 69.7 \\
& \textcolor{black}{TruMPET} & 0.672 & 0.672 & 0.672 & 0.658 & 0.594 & 66.6 \\
& \textbf{MOGP-MMF (ours)} & \textbf{0.690} & \textbf{0.672} & \textbf{0.690} & \textbf{0.668} & \textbf{0.595} & \textbf{70.4} \\
\bottomrule
\end{tabular*}
\end{table*}

\begin{table*}[htbp]
\caption{Performance Comparison of MMF and Alternative Feature Configurations (Accuracy in Percentage)}
\label{MVMF}
\centering
\renewcommand{\arraystretch}{0.98}%
\setlength{\tabcolsep}{2mm}{%
\begin{tabular}{@{}l|l|lllllllll@{}}
\toprule
Method & Representation &CB6133 &CB6133 &CB513 &CASP10 &CASP11& CASP12& CASP13& CASP14
\\
&&(Validation)&(Test)&&&\\
\midrule
\multirow{16}{*}{Single-view Single-level feature} & HMM-CNN1 & 69.86 & 70.35 & 66.91 & 69.61 & 68.14 &66.47&63.46&61.50\\
 & HMM-CNN2 & 72.11 & 72.64 & 69.33 & 71.39 & 70.50 &46.69&45.70&42.87\\
 & HMM-RNN1 & 72.58 & 72.99 & 69.56 & 70.95 & 70.59 &69.19&66.80&65.33\\
 & HMM-RNN2 & 72.46 & 72.91 & 69.37 & 71.84 & 70.54 &69.68&66.29&65.81\\
 & PSSM-CNN1 & 69.44 & 69.58 & 66.33 & 68.91 & 66.24 &59.70&58.63&59.53\\
 & PSSM-CNN2 & 71.41 & 71.87 & 68.85 & 71.78 & 69.15 &64.62&63.38&62.88\\
 & PSSM-RNN1 & 72.71 & 72.86 & 69.84 & 72.68 & 70.41 &66.37&64.20&62.85\\
 & PSSM-RNN2 & 71.29 & 72.90 & 70.46 & 73.49 & 71.05 &66.48&63.88&62.97\\
 & T5-CNN1 & 78.69 & 78.77 & 75.27 & 79.36 & 70.06 &48.51&47.30&45.45\\
 & T5-CNN2 & 78.76 & 79.24 & 75.44 & 79.49 & 75.92 &73.52&71.65&68.02\\
 & T5-RNN1 & 78.36 & 78.48 & 74.99 & 78.90 & 75.52 &72.46&70.77&67.03\\
 & T5-RNN2 & 78.50 & 78.55 & 74.55 & 78.90 & 75.52 &73.33&71.48&67.53\\
 & Sa-CNN1 & 69.71 & 70.17 & 65.66 & 71.73 & 68.37 &65.71&63.01&62.75\\
 & Sa-CNN2 & 70.05 & 70.53 & 66.62 & 71.76 & 68.51 &66.39&62.96&62.41\\
 & Sa-RNN1 & 69.72 & 70.06 & 65.69 & 71.83 & 68.35 &66.07&62.25&61.88\\
 & Sa-RNN2 & 69.57 & 70.02 & 65.12 & 71.11 & 68.37 &65.95&62.95&62.32\\
\midrule
\multirow{4}{*}{Single-view Multi-level feature} & HMM & 73.40 & 73.56 & 72.52 & 71.56 & 71.75 &69.41&67.11&66.13\\
 & PSSM & 73.96 & 73.87 & 71.21 & 74.11 & 71.75 &67.21&65.06&63.68\\
 & T5 & 79.36 & 79.45 & 76.03 & 79.98 & 75.10 &74.09&72.64&68.65\\
 & Sa & 70.63 & 70.98 & 66.64 & 72.60 & 69.06 &66.63&63.08&62.81\\
\midrule
\multirow{4}{*}{Multi-view Single-level feature} & CNN1 & 79.10 & 79.07 & 75.66 & 79.49 & 76.22 &72.90&71.24&68.06\\
 & CNN2 & 79.37 & 79.53 & 76.02 & 79.97 & 76.36 &73.63&71.54&68.63\\
 & RNN1 & 79.06 & 79.03 & 75.85 & 79.46 & 76.44 &74.06&71.65&68.15\\
 & RNN2 & 79.19 & 79.19 & 75.67 & 79.46 & 76.44 &73.98&71.89&68.77\\
\midrule
MOGP-MMF & Ours & \textbf{80.15} & \textbf{80.01} & \textbf{76.96} & \textbf{80.49} & \textbf{77.58} &\textbf{74.60} & \textbf{72.11} & \textbf{69.02}\\
\bottomrule
\end{tabular}%
}\
\end{table*}

\subsection{Comparison with the state-of-the-art methods}
\textcolor{black}{We benchmarked MOGP-MMF against nine expert-driven models, including classic baselines (e.g., NetSurfP-3.0, AttSec) and recent SOTA approaches from 2025 (e.g., TruMPET, DeepPredict) across seven datasets using six diverse metrics, as detailed in Table \ref{tab:performance_comparison}.}

\textcolor{black}{MOGP-MMF achieves the highest Q8 accuracy across all test sets, notably outperforming recent competitors. On CASP10 and CASP11, it attains 80.5\% and 77.6\%, surpassing the nearest baselines (DeepPredict and TruMPET) by margins of 1.2\% and 1.0\%, respectively. Even on the low-homology CASP14, it maintains robust performance (69.0\%), validating its adaptability to difficult targets.}

\textcolor{black}{Crucially, MOGP-MMF excels in Segment Overlap (Sov), outperforming the runner-up by 2.2\% on CASP12. This indicates superior recovery of continuous structural segments rather than fragmented residues. Furthermore, it achieves the leading F1-score (0.793) and MCC on CB6133, demonstrating an effective balance between Precision and Recall despite class imbalance.}

\textcolor{black}{In summary, the success of MOGP-MMF highlights the efficacy of its automated evolutionary framework and multi-view feature fusion in adaptively capturing complex sequence-structure dependencies.}


\subsection{Analysis of MMF method}

To evaluate the effectiveness of the MMF strategy, we compared it with other feature configuration approaches under the MOGP framework, including single-view single-level, single-view multi-level, and multi-view single-level feature configurations. Table \ref{MVMF} shows that MMF consistently achieves superior performance across all datasets.

The results indicate that single-view single-level feature models yield the lowest accuracy, suggesting that relying on a single feature source is insufficient to capture the complexity of protein sequences. Although single-view multi-level configurations outperform single-view single-level counterparts, they still underperform compared to multi-view approaches. This underscores the importance of incorporating diverse feature views, such as sequence information, evolutionary data, and contextual relationships. While multi-view single-level models show some advantages, they do not match the performance of MMF, highlighting its ability to provide richer feature representations and better capture the intricate relationships between protein sequences and structures.

\subsection{Evaluation of EOS}
\begin{figure*}[htbp]
    \centering
    \begin{subfigure}[b]{0.48\textwidth} 
        \centering
        \includegraphics[width=\textwidth]{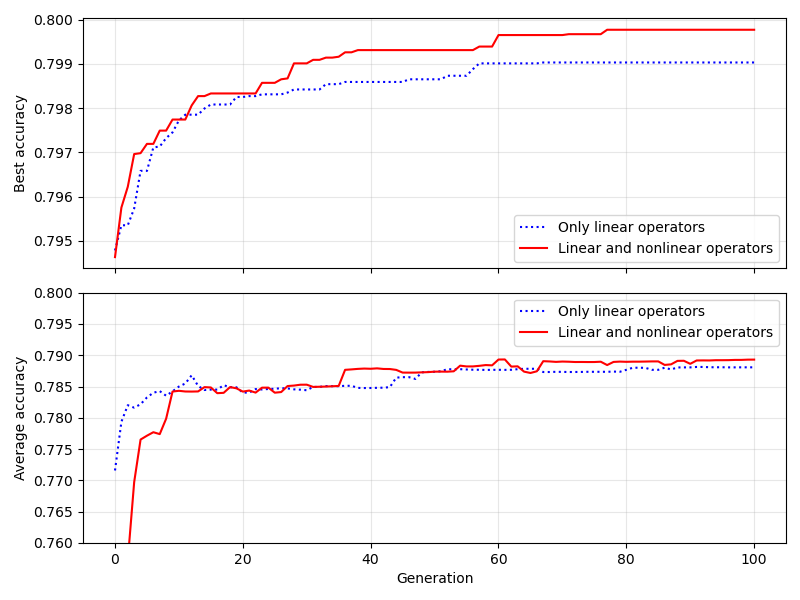}
        \caption{Accuracy comparison under MOGP}
        \label{EOS1}
    \end{subfigure}
    \hspace{0.05cm}
    \begin{subfigure}[b]{0.48\textwidth}
        \centering
        \includegraphics[width=\textwidth]{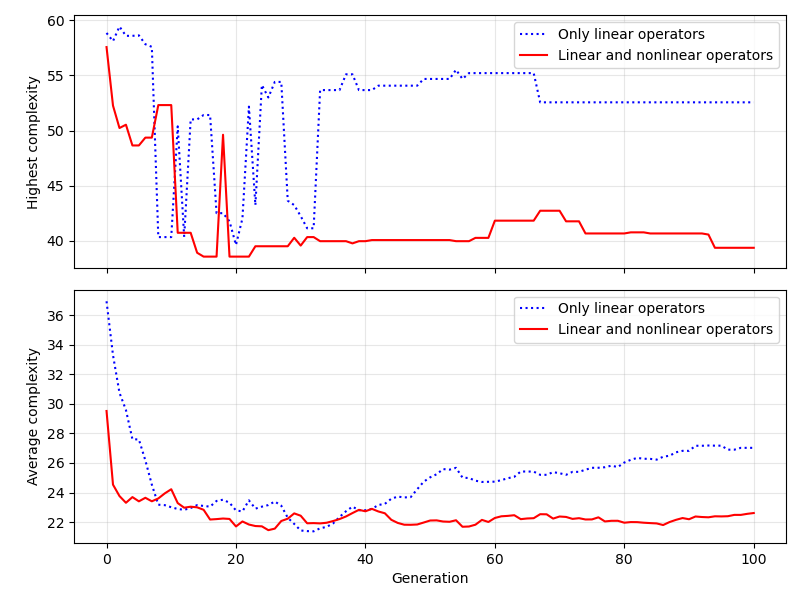}
        \caption{Complexity comparison under MOGP}
        \label{EOS2}
    \end{subfigure}

    \caption{\small Visualization of comparisons between models \textit{with} and \textit{without} nonlinear operators on CB6133 dataset: (a) Accuracy comparison under MOGP, (b) Complexity comparison under MOGP.}
    \label{EOS}
\end{figure*}

\textcolor{black}{To validate the effectiveness of the proposed EOS, specifically the contribution of nonlinear operators, we compared the Linear and nonlinear operators set against a Linear-only baseline. Figures \ref{EOS1} and \ref{EOS2} illustrate the evolutionary dynamics of model accuracy and structural complexity, respectively.}

\textcolor{black}{As shown in Figure \ref{EOS1}, both strategies converge similarly in early stages. However, the inclusion of nonlinear operators (red line) enables the population to break through performance bottlenecks around the 30th generation, reaching a higher plateau. While the linear baseline struggles with local optima, the nonlinear EOS yields superior and more stable average and best accuracy during late-stage evolution.}

\textcolor{black}{Crucially, Figure \ref{EOS2} demonstrates that EOS significantly reduces model complexity. The linear-only approach suffers from severe "operator bloat," with maximum complexity oscillating between 50 and 60. In contrast, nonlinear operators capture complex biological patterns more efficiently, effectively constraining complexity to approximately 40. This confirms that nonlinear functions provide a more expressive vocabulary, allowing EOS to achieve higher accuracy with simpler program structures, thereby mitigating bloat while ensuring high performance.}


\begin{table*}[htbp]
\caption{Performance Comparison of MOGP and Alternative Fusion Methods (Accuracy in Percentage)}
\label{fusion_methods}
\centering
\renewcommand{\arraystretch}{0.9}%
\setlength{\tabcolsep}{2mm}{%
\begin{tabular}{@{}lllllllll@{}}
\toprule
Method & CB6133  & CB6133 & CB513 & CASP10 & CASP11&CASP12&CASP13&CASP14 \\
& (Validation)&(Test) &&&&&\\
\midrule
Add & 78.00 & 78.47 & 74.89 & 78.55 & 75.85 & 73.60& 71.39& 68.35\\
Mul & 54.61 & 55.75 & 50.70 & 52.11 & 54.50 & 50.51& 47.49& 47.41\\
Max & 77.97 & 78.25 & 74.75 & 78.55 & 75.62 & 72.08& 69.59& 67.79\\
Min & 74.00 & 74.17 & 71.09 & 74.30 & 71.98 & 70.25& 68.46& 66.32\\
Concatenation & 74.88 & 75.38 & 71.97 & 75.62 & 73.34& 70.74&67.44 &66.59\\
\midrule
\textbf{MOGP-MMF} & \textbf{80.15} & \textbf{80.01} & \textbf{76.96} & \textbf{80.49} & \textbf{77.58} &\textbf{74.60} & \textbf{72.11} & \textbf{69.02}\\
\bottomrule
\end{tabular}%
}\
\end{table*}

\begin{figure*}[htbp]
    \centering
    \begin{subfigure}[b]{0.40\textwidth} 
        \centering
        \includegraphics[width=\textwidth]{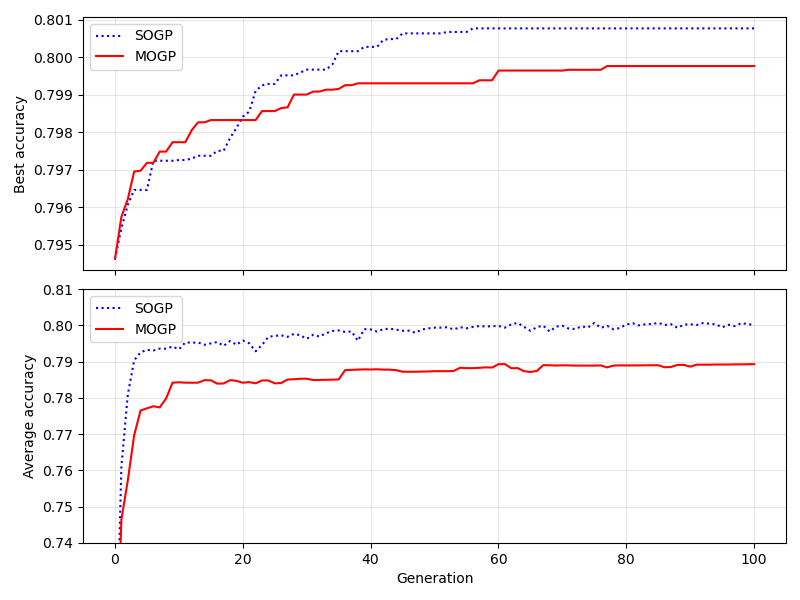}
        \caption{\small Accuracy comparison between SOGP and MOGP}
        \label{k1}
    \end{subfigure}
    \hspace{0.05cm}
    \begin{subfigure}[b]{0.40\textwidth}
        \centering
        \includegraphics[width=\textwidth]{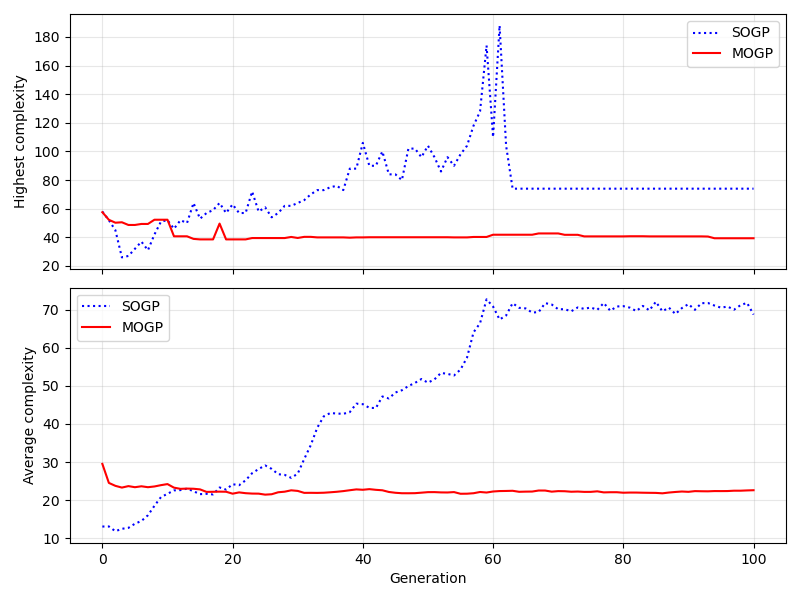}
        \caption{\small Complexity comparison between SOGP and MOGP}
        \label{k2}
    \end{subfigure}
\vspace{0.05cm}

    \begin{subfigure}[b]{0.40\textwidth}
        \centering
        \includegraphics[width=\textwidth]{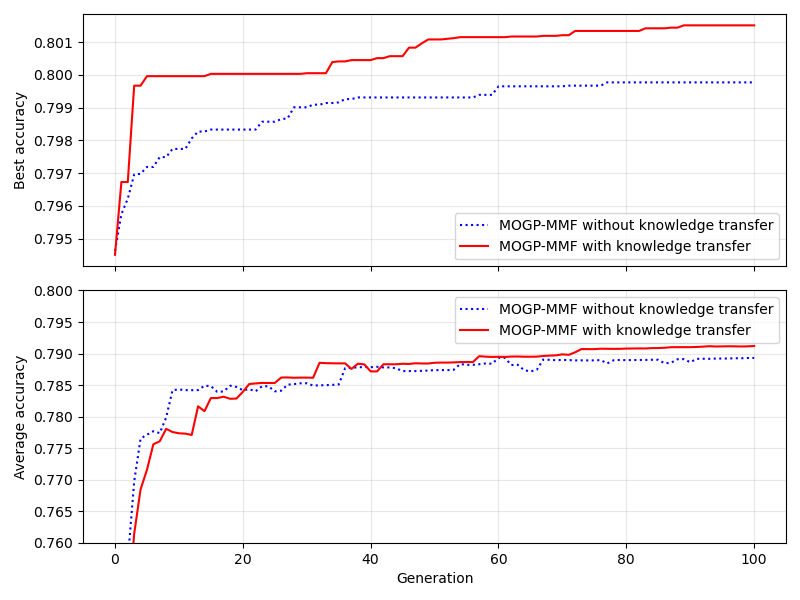}
        \caption{\small MOGP Accuracy with vs. without KT}
        \label{k3}
    \end{subfigure}
         \hspace{0.05cm}
    \begin{subfigure}[b]{0.40\textwidth}
        \centering
        \includegraphics[width=\textwidth]{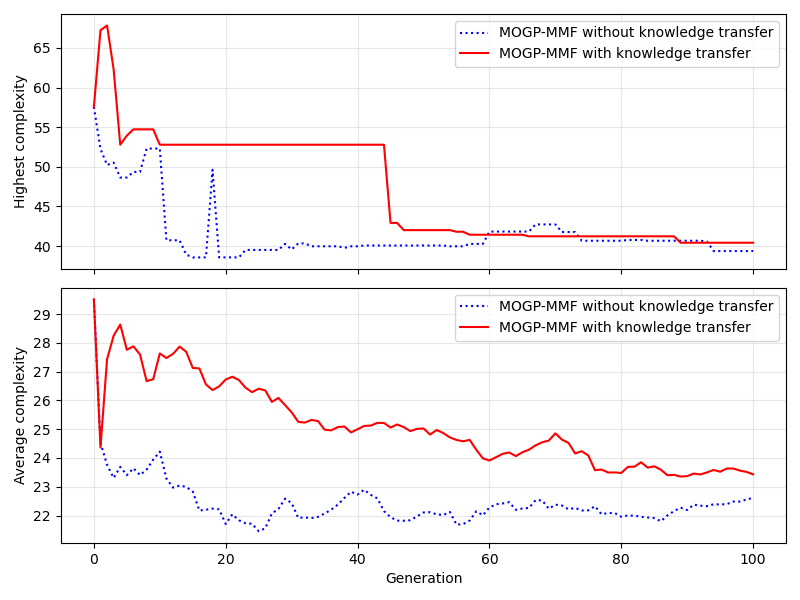}
        \caption{\small MOGP Complexity with vs. without KT}
        \label{k4}
    \end{subfigure}
    
    \caption{\small Performance comparisons on the CB6133 dataset. (a-b) Accuracy and complexity of SOGP vs. Naive MOGP. (c-d) Impact of Knowledge Transfer (KT) on MOGP accuracy and complexity.}
    \label{KT}
\end{figure*}

\subsection{Analysis of the MOGP fusion mechanism}

To evaluate fusion efficacy, we benchmarked MOGP against five manual strategies: Add, Max, Min, Mul, and Concatenation (Table \ref{fusion_methods}). MOGP-MMF consistently outperforms all fixed baselines.

Analysis of the results reveals the limitations of traditional methods. While Add and Max yield competitive results by capturing dominant signals, their static nature prevents modeling complex, non-linear inter-view interactions. Notably, Concatenation—the deep learning standard—underperforms, suggesting that high-dimensional stacking often introduces redundancy rather than synergy. Conversely, Mul and Min suffer from severe "signal vanishing".

In contrast, MOGP's superiority stems from its evolutionary capability to autonomously generate adaptive, non-linear expressions. This approach effectively exploits complementary information across views while filtering out noise, ensuring exceptional robustness and generalization where rigid, predefined strategies often fall short.


\subsection{Analysis of knowledge transfer method in MOGP}

\textcolor{black}{To evaluate the efficacy of the knowledge transfer (KT) strategy, we conducted a comparative analysis of the evolutionary dynamics across Single-Objective (SO), Naive Multi-Objective (MO), and MOGP-MMF, as illustrated in Fig. \ref{KT}. The experimental results highlight a performance trade-off challenge commonly encountered in most conventional standard optimization strategies.}

\textcolor{black}{Specifically, while the SO approach achieves high predictive accuracy, Fig. \ref{k2} indicates a tendency toward elevated structural complexity. This arises as the algorithm often generates deeper or more redundant tree structures in an effort to maximize training scores. In contrast, although the Naive MO method effectively constrains complexity (as shown in Fig. \ref{k2}), this constraint appears to restrict the search space. The results in Fig. \ref{k1} suggest that this emphasis on parsimony can lead the population into local optima, resulting in a discernible "accuracy degradation."}

\textcolor{black}{The KT mechanism acts as a crucial bridge to reconcile this trade-off. As illustrated in Fig. \ref{k3}, by leveraging high-performing SO individuals as prior knowledge, MOGP-MMF restores predictive accuracy to levels comparable with the SO baseline. This enhancement is achieved while preserving the low model complexity characteristic of the MO framework, a trend further confirmed by the results in Fig. \ref{k4}. Consequently, KT is essential to mitigate limitations inherent in naive multi-objective approaches.}

\begin{figure*}[htbp]
  \centering
  \includegraphics[width=18cm,height=6cm]{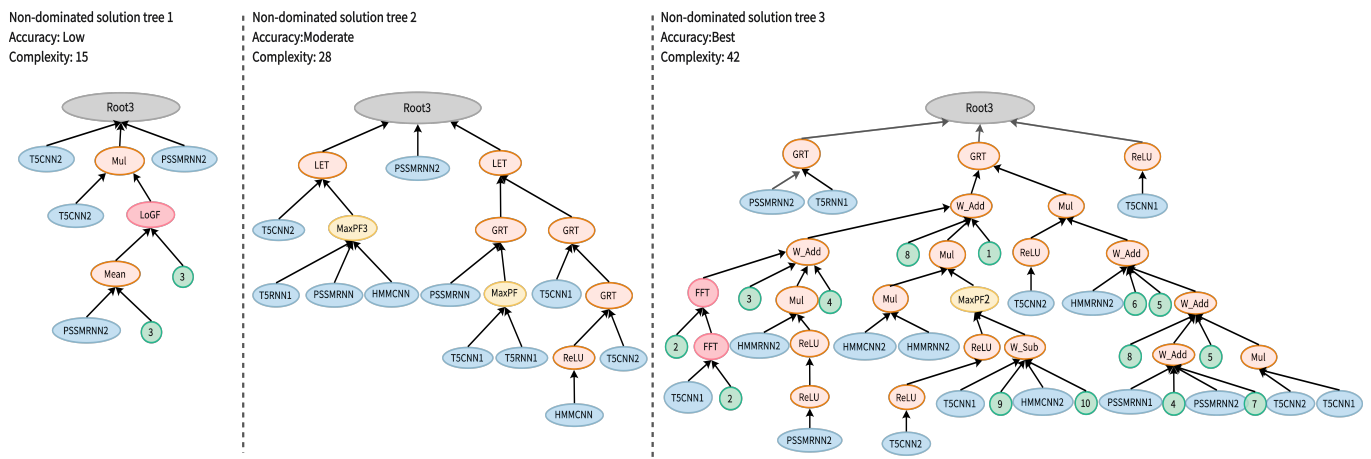}
  \caption{\small \textcolor{black}{Representative "anchor" trees from the 200-solution Pareto front (CB6133): (a) Low-Complexity, (b) Balanced, and (c) High-Performance.}}
  \label{solution}
\end{figure*}

\begin{table*}[htbp]
\caption{\textcolor{black}{\small Representative "anchor" solution trees selected from the 200 non-dominated individuals in the 100th generation (CB6133 dataset): (a) Tree 1 (Low-Complexity), (b) Tree 2 (Balanced), and (c) Tree 3 (High-Performance). (Accuracy in Percentage)}}
\label{trees}
\centering
\renewcommand{\arraystretch}{1}%
\setlength{\tabcolsep}{2mm}{%
\begin{tabular}{@{}llllllllll@{}}
\toprule
Method   &Anchor &Complexity & CB6133 (Test) & CB513 & CASP10 & CASP11 &CASP12&CASP13&CASP14\\
\midrule
Tree 1  &Low-Complexity&15 &79.39 & 75.98 & 79.76 & 76.41 &68.28&66.15&62.70\\
Tree 2  &Balanced&28&79.91 & 76.89 & 80.32 & 77.26 &73.16&71.19&67.50\\
Tree 3 &High-Performance&42 & 80.05 & 76.93 & 80.47 & 77.59 &74.63&72.09&69.07\\
\bottomrule
\end{tabular}%
}\
\end{table*}


\subsection{Multiple outperforming solutions for flexible choices}

\textcolor{black}{MOGP-MMF generates a dense Pareto front of non-dominated solutions via multi-objective optimization mechanism. To demonstrate the deployment flexibility, we analyze three representative "anchor" solutions selected from the final population of 200 individuals in the 100th generation trained on CB6133, as illustrated in Fig.~\ref{solution}.}

These three solutions span diverse application scenarios: Tree 1 (Low-Complexity Anchor), with a complexity of only 15, represents the lightweight deployment scenario, where minimal computational overhead is prioritized despite a slight compromise in accuracy; Tree 2 (Balanced Anchor), located at the "knee point" of the Pareto front with a moderate complexity of 28, offers an optimal trade-off between performance and efficiency; and Tree 3 (High-Performance Anchor) demonstrates the potential for maximum performance when complexity constraints are relaxed, ultimately reaching a higher structural complexity of 42.


\textcolor{black}{As detailed in Table \ref{trees}, higher complexity enables the adaptive fusion of a broader spectrum of views and features. This deep feature integration enables Tree 3 to excel in handling challenging samples; specifically, it achieves significantly higher accuracy on the CASP12–CASP14 datasets compared to lower-complexity models. In conclusion, MOGP-MMF exhibits high diversity and adaptability, providing users with a rich pool of high-quality candidate solutions tailored to specific computational constraints.}

\begin{figure*}[p] 
    \centering
    \setlength{\tabcolsep}{1pt} 

    \makebox[0.32\textwidth]{\small \textbf{(a) Single-feature}} \hfill
    \makebox[0.32\textwidth]{\small \textbf{(b) MMF}} \hfill
    \makebox[0.32\textwidth]{\small \textbf{(c) MOGP-MMF}} \\
    \vspace{0cm} 



    \includegraphics[width=0.32\textwidth]{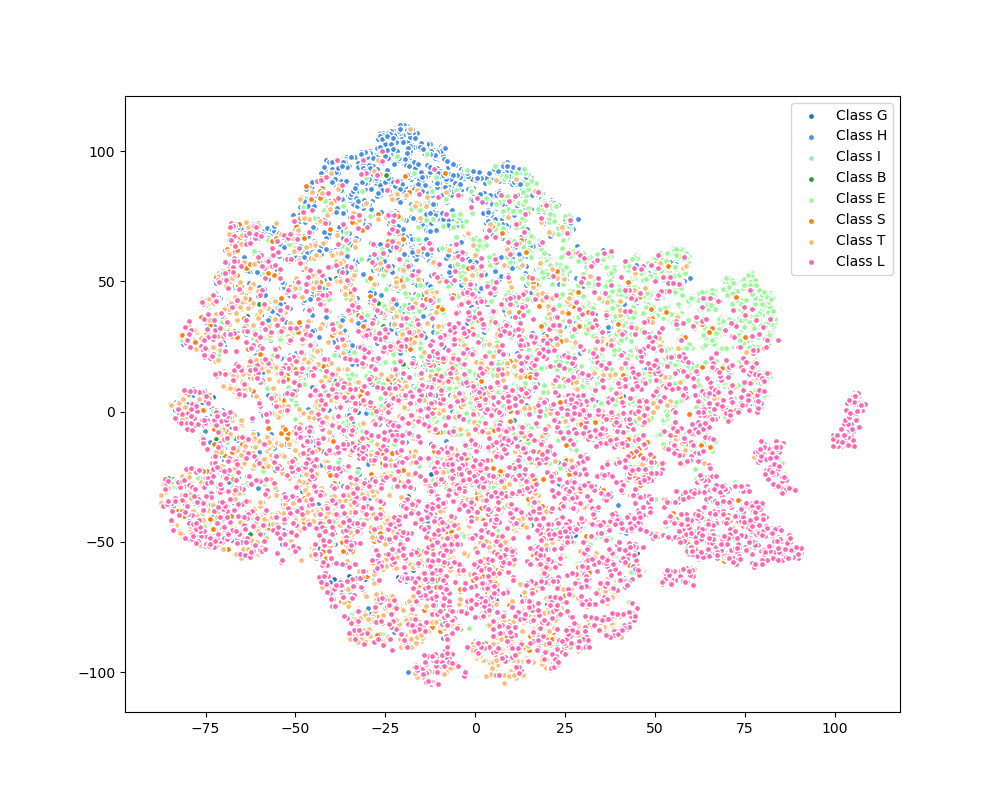} \hfill
    \includegraphics[width=0.32\textwidth]{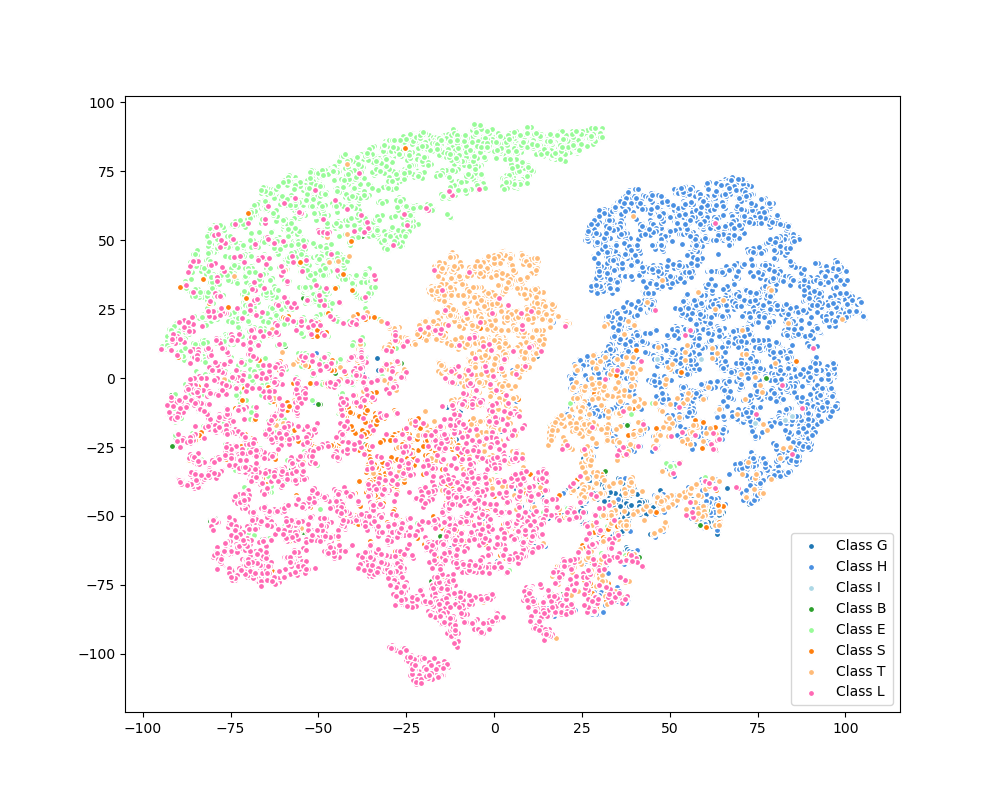} \hfill
    \includegraphics[width=0.32\textwidth]{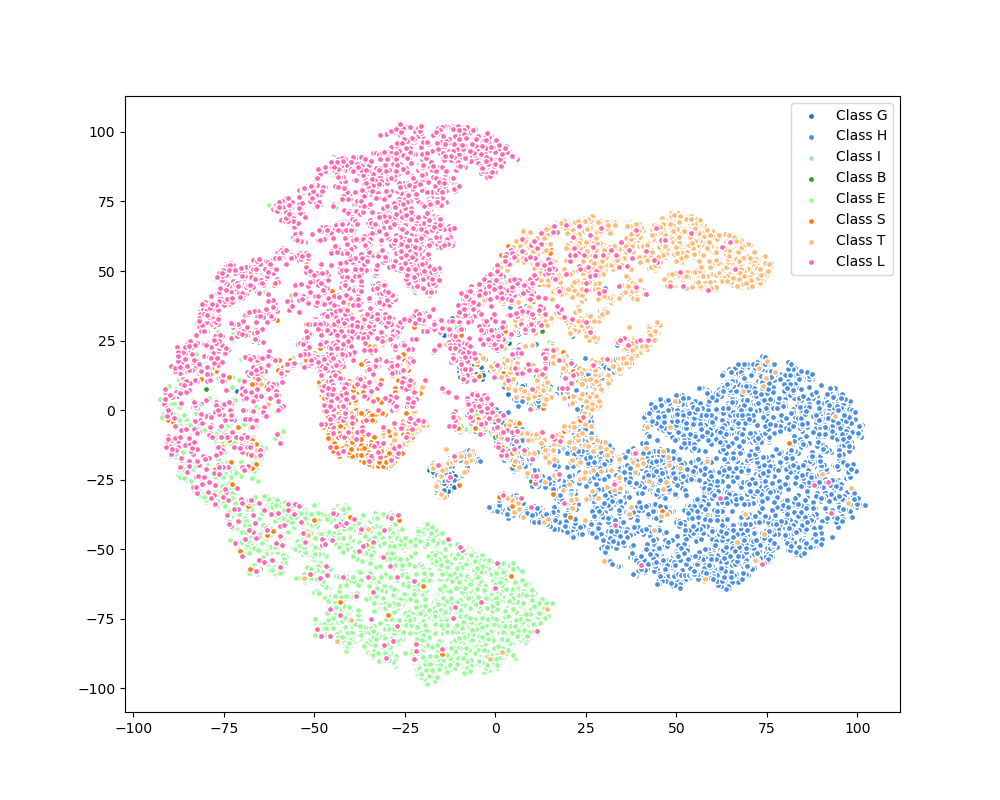} \\
    \vspace{-0.4cm}

    \includegraphics[width=0.32\textwidth]{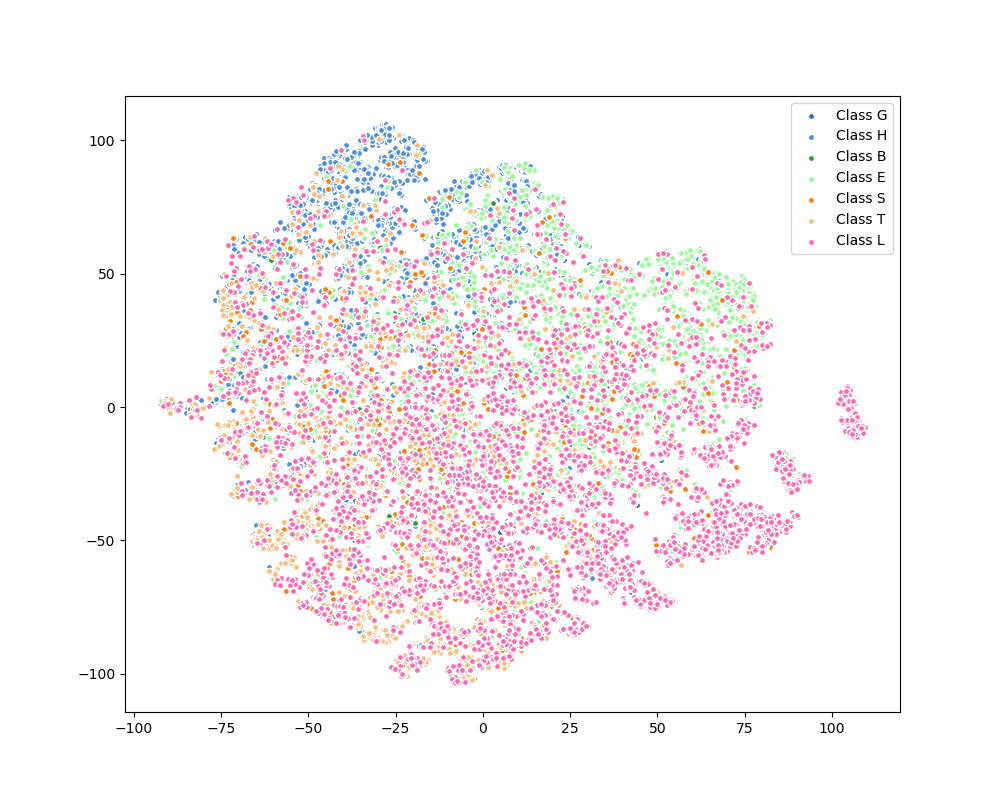} \hfill
    \includegraphics[width=0.32\textwidth]{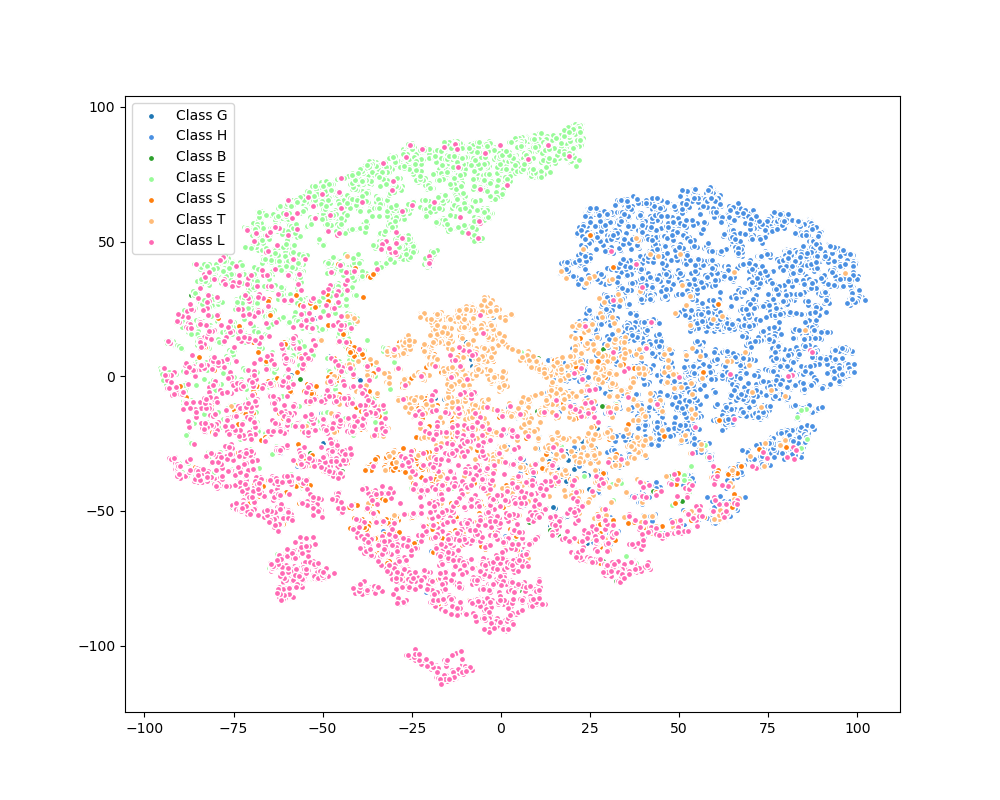} \hfill
    \includegraphics[width=0.32\textwidth]{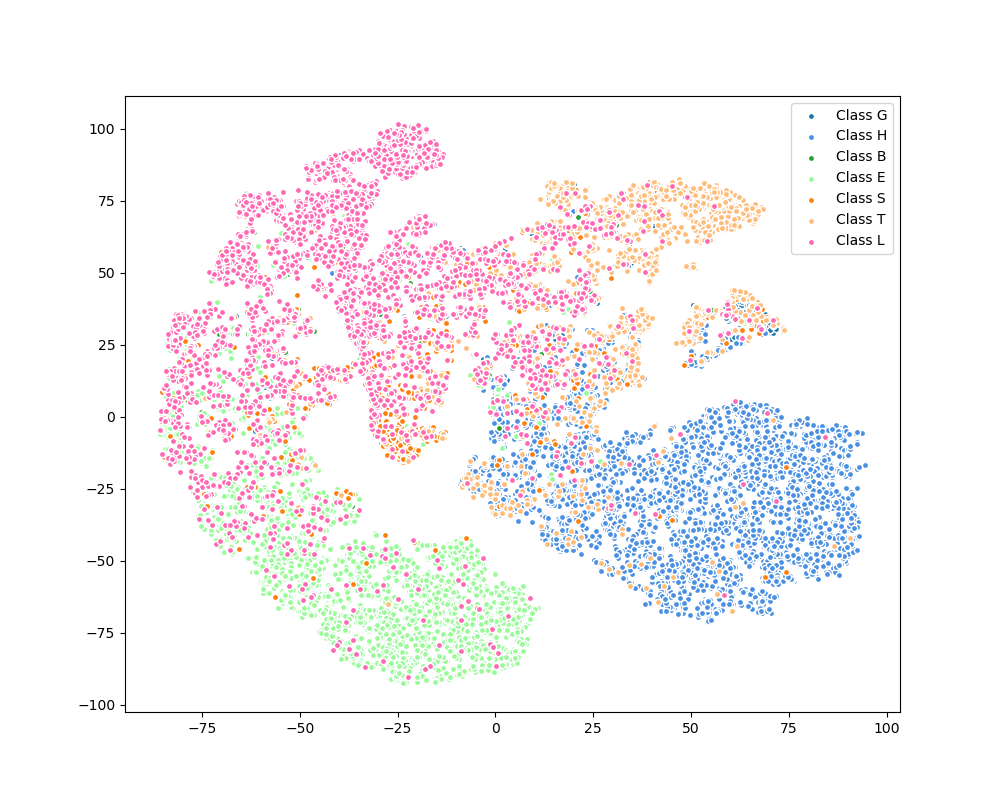} \\
    \vspace{-0.4cm}

    \includegraphics[width=0.32\textwidth]{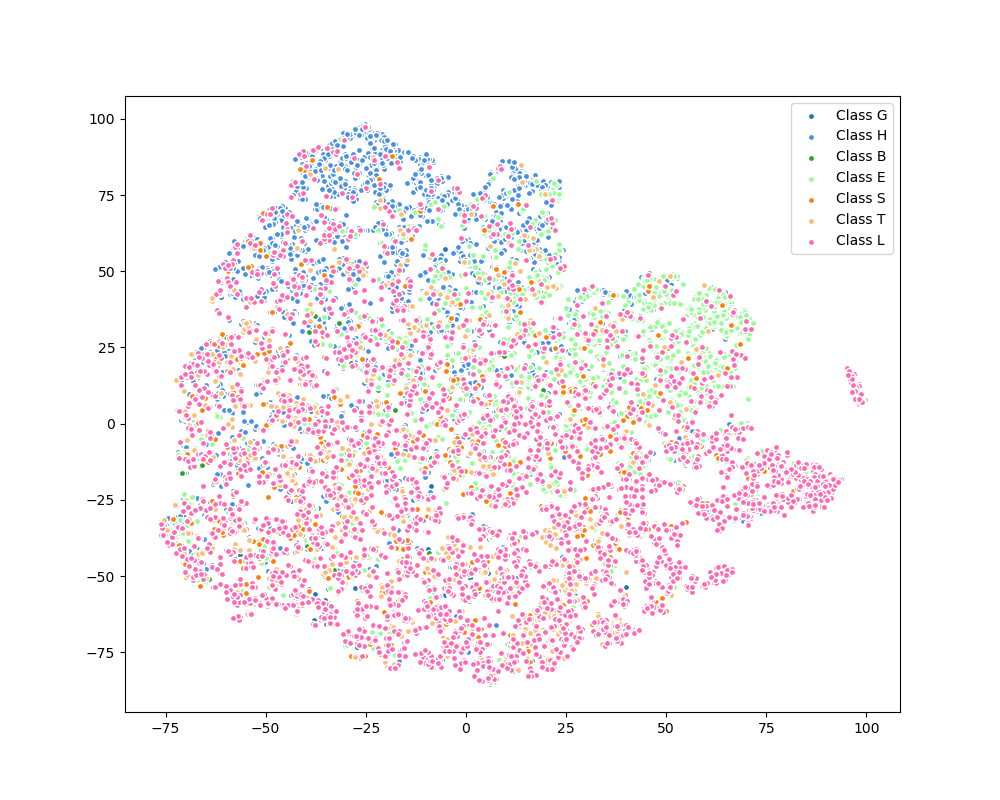} \hfill
    \includegraphics[width=0.32\textwidth]{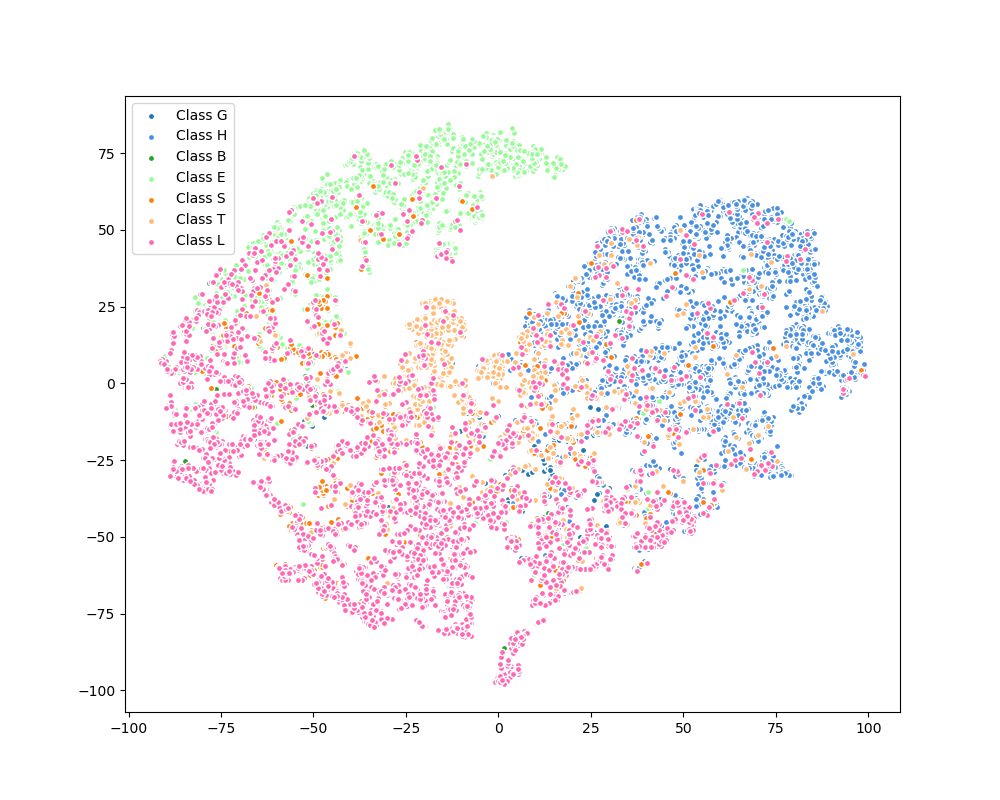} \hfill
    \includegraphics[width=0.32\textwidth]{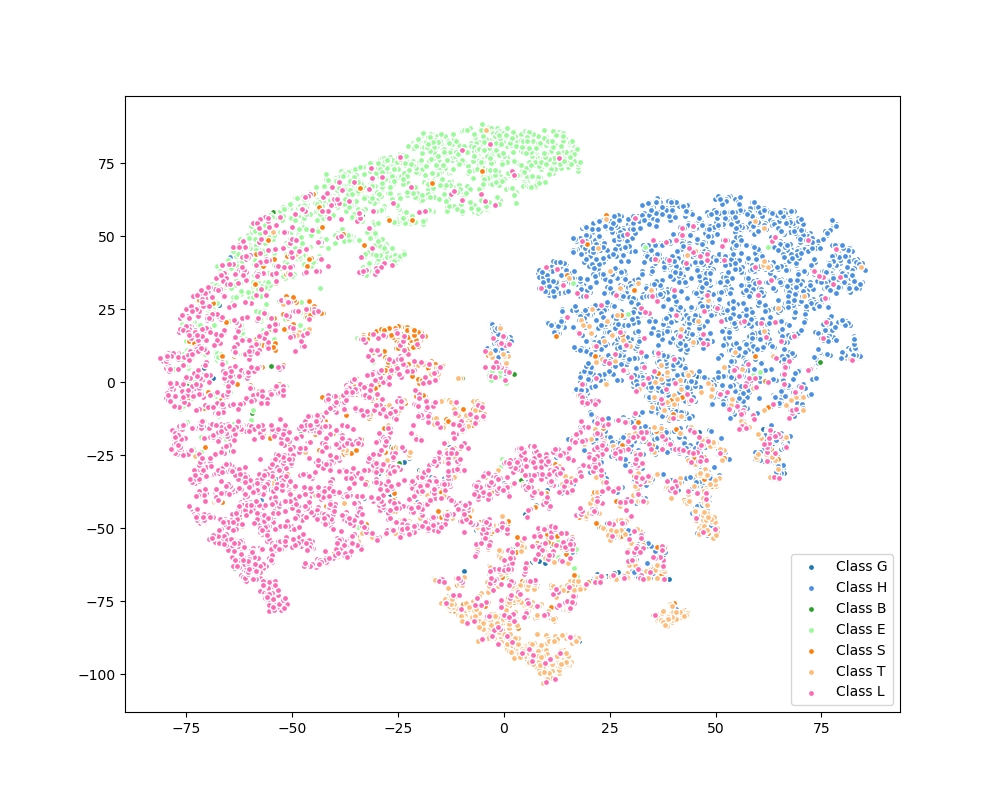} \\
    \vspace{-0.4cm}

    \includegraphics[width=0.32\textwidth]{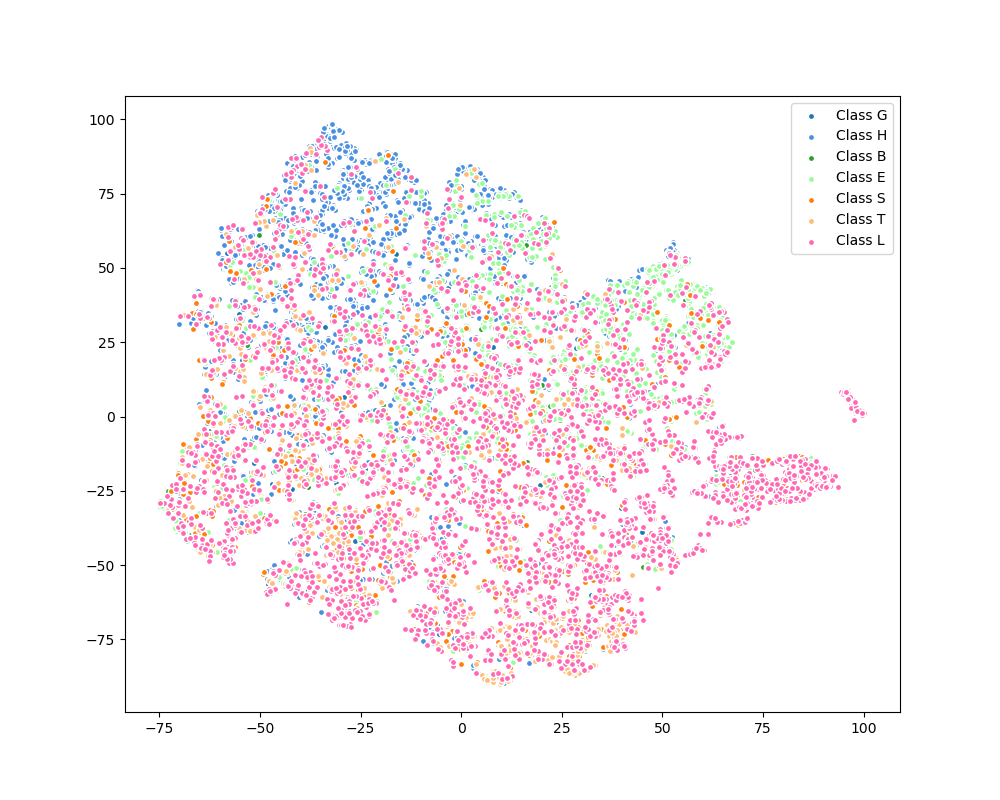} \hfill
    \includegraphics[width=0.32\textwidth]{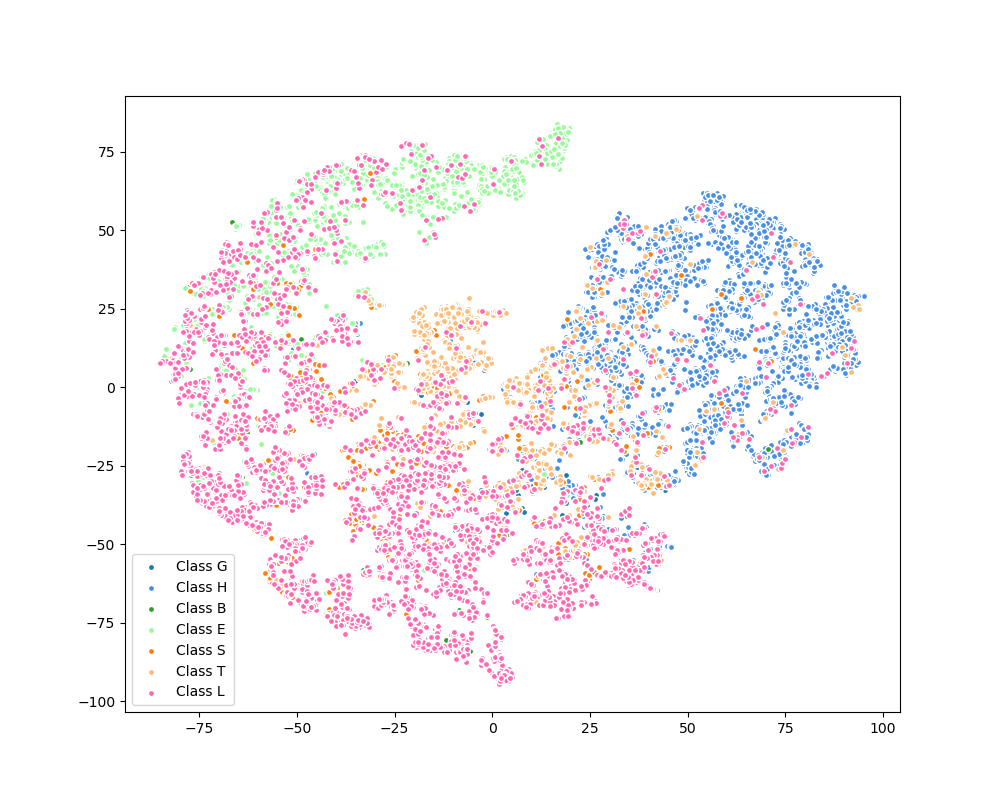} \hfill
    \includegraphics[width=0.32\textwidth]{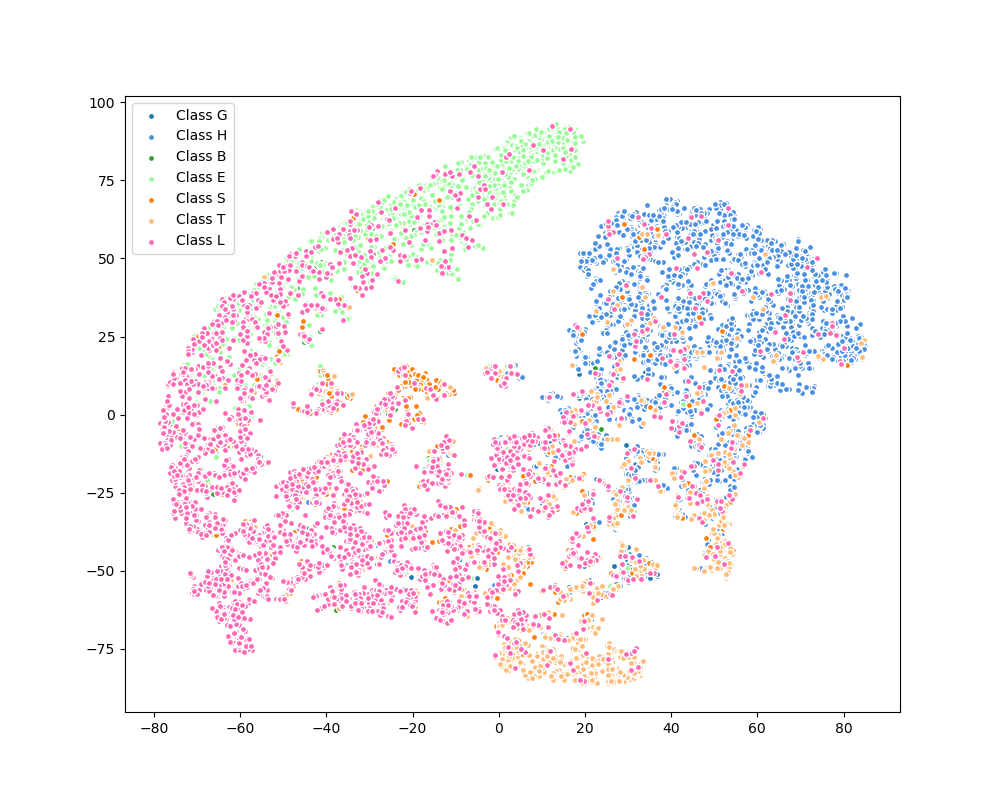} \\
    \vspace{-0.4cm}

    \includegraphics[width=0.32\textwidth]{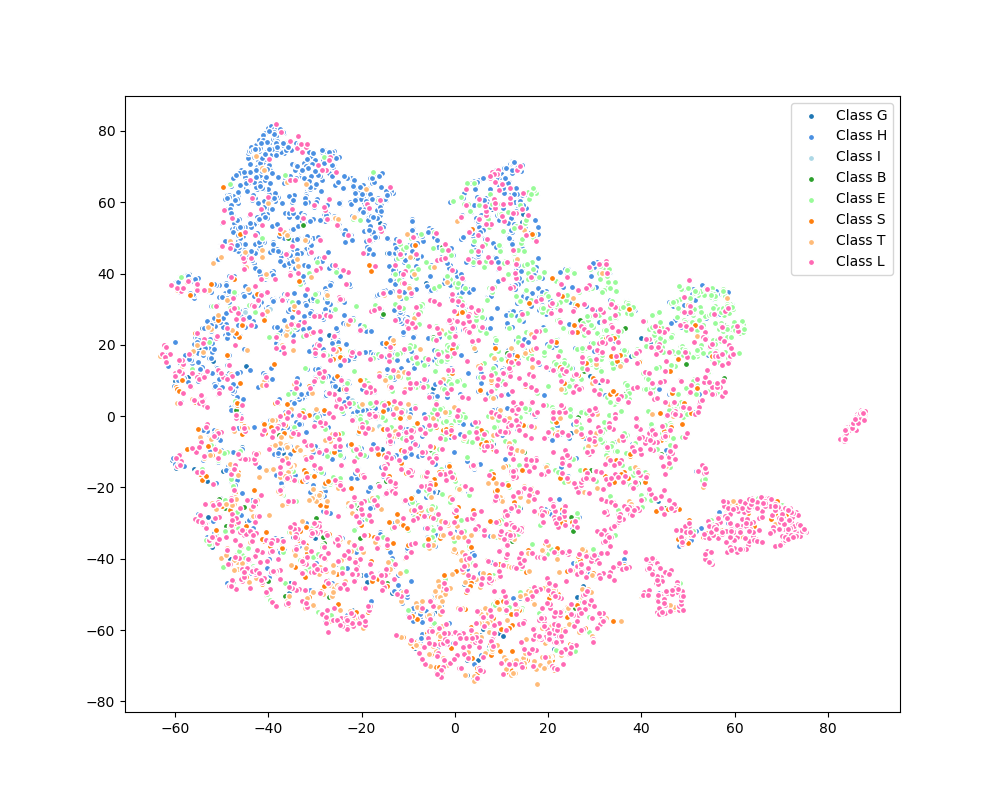} \hfill
    \includegraphics[width=0.32\textwidth]{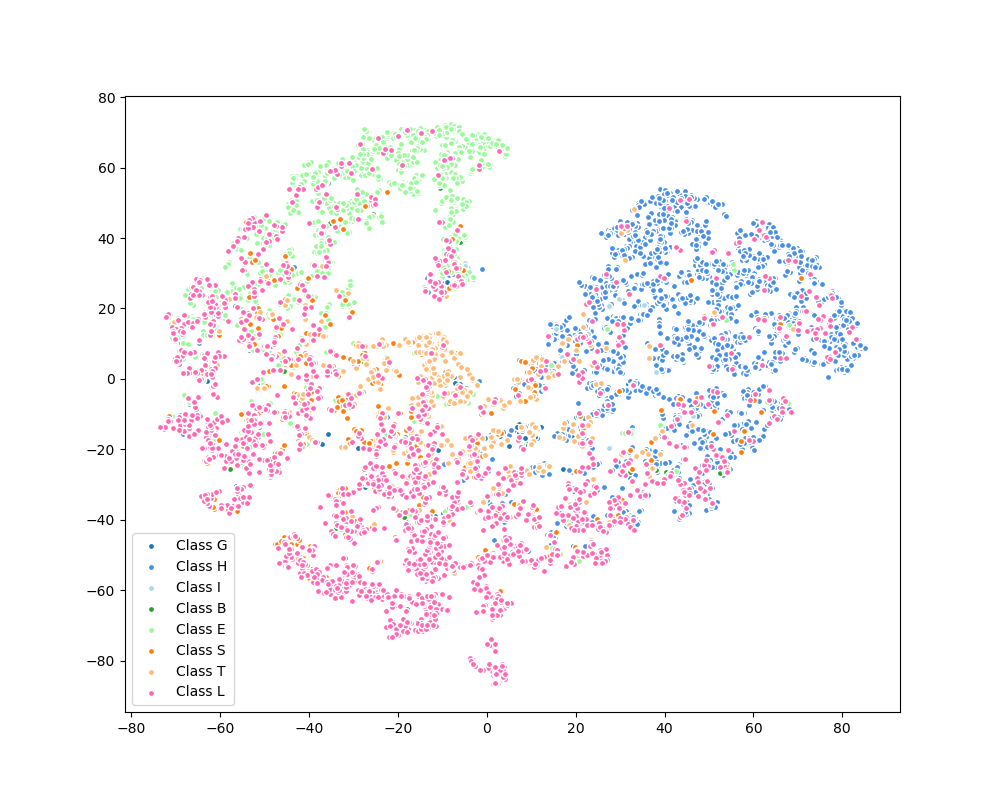} \hfill
    \includegraphics[width=0.32\textwidth]{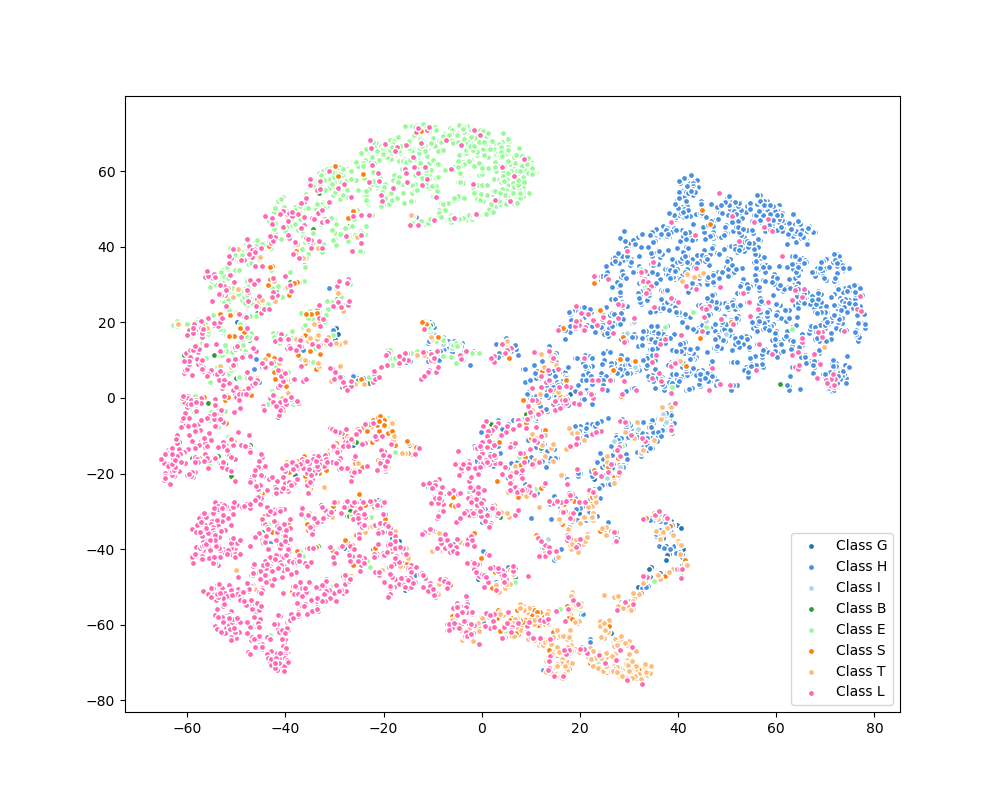} \\
    
    \vspace{-0.4cm} 
    
    \caption{\small Visualization of feature representations across seven datasets. \textbf{Rows (top to bottom):} CASP10, CASP11, CASP12, CASP13, and CASP14. \textbf{Columns:} (a) Single-feature baseline, (b) Standard Multi-View Multi-Level Fusion (MMF), and (c) The proposed MOGP-MMF framework.}
    \label{combined_figure}
\end{figure*}

\subsection{Visual analysis of feature representations}

This section presents a visualization analysis of how MOGP-MMF transforms hidden features using t-SNE \cite{visual}, a widely adopted non-linear dimensionality reduction technique. We benchmarked the feature space quality across the CASP10 to CASP14 datasets (corresponding to rows 1--5 in the figure). The comparison involves three methods: the single-feature method using CNN1-extracted one-hot encodings (Left Column), the MMF method utilizing the static 'Add' strategy (Middle Column), and our MOGP-MMF using the optimal evolutionary tree (Right Column). For clarity, data points are colored by their ground-truth secondary structure labels: Helices (G, H, I) are shown in shades of blue, Strands (B, E) in shades of green, and Coils (S, T, L) in shades of orange-pink.

Observations from the Left Column reveal poor clustering performance with significant class overlap, suggesting that single-feature representations struggle to effectively differentiate complex protein structures. The Middle Column (MMF) exhibits improved clustering and reduced overlap; however, class boundaries remain indistinct, indicating that while static multi-view fusion offers gains over single features, it lacks optimal integration. Conversely, the Right Column (MOGP-MMF) achieves the best clustering performance, exhibiting sharp class boundaries and minimal overlap. Although minor misclassifications persist, MOGP-MMF significantly outperforms both baseline methods in generating distinct and robust feature representations, which directly translates to its enhanced prediction accuracy.

\section{Discussion}
\textcolor{black}{Based on the above evaluations, we conclude that MOGP-MMF offers three distinct advantages: (1) \textbf{Superior Performance and Robustness:} MOGP-MMF achieves SOTA results across seven benchmarks and six metrics. Its dominance in Q8, MCC, and Sov specifically validates its advanced predictive power and generalization capability across diverse structural classes. (2) \textbf{Adaptive Fusion:} Unlike traditional methods relying on static concatenation—which often introduce noise and redundancy—our evolutionary mechanism autonomously selects and fuses discriminative features. This effectively mitigates the "curse of dimensionality" and ensures robust representations. (3) \textbf{Flexibility:} By employing multi-objective optimization, MOGP-MMF generates a Pareto front of non-dominated solutions rather than a single fixed model. This enables trade-offs between accuracy and complexity, allowing users to select lightweight models for edge devices or high-precision variants for server-side analysis.}

\textcolor{black}{However, two limitations warrant comparison: (1) Training Overhead: The evolutionary process is computationally more intensive than gradient-based training. Yet, this cost is strictly offline; online inference remains exceptionally fast due to the compact nature of the evolved formulas. (2) Feature Dependency: Performance is partially bounded by the quality of initial views. Future integration of advanced protein language models and structural embeddings could further elevate the predictive ceiling.}

\section{CONCLUSION}

In this paper, we reformulate the protein secondary structure prediction problem as an optimization task focused on feature selection and fusion, proposing a multi-objective genetic programming-based solution (MOGP-MMF). The proposed method utilizes deep learning for feature extraction and employs a multi-view, multi-feature strategy to enrich protein representations. Additionally, an enriched operator set, comprising both linear and nonlinear operators, enhances model accuracy while reducing complexity. The incorporation of knowledge transfer techniques further improves performance and applicability. Experimental results show that MOGP-MMF enhances model performance, surpassing state-of-the-art methods across key metrics while eliminating dependence on expert-designed features. Moreover, MOGP-MMF provides multiple non-dominated solutions to accommodate diverse application needs and offers visualization of the learned features.






\ifCLASSOPTIONcaptionsoff
  \newpage
\fi

\end{CJK*}

\end{document}